\documentclass{article}

\usepackage[final,nonatbib]{neurips_2022}
\usepackage[square,sort,numbers]{natbib}

\usepackage[utf8]{inputenc} 
\usepackage[T1]{fontenc}    
\usepackage{hyperref}       
\usepackage{url}            
\usepackage{booktabs}       
\usepackage{amsfonts}       
\usepackage{nicefrac}       
\usepackage{microtype}      
\usepackage{xcolor}         

\usepackage{amsmath}
\usepackage{anyfontsize}
\usepackage{caption}
\usepackage{enumitem}
\usepackage{float}
\usepackage{rotating}
\usepackage{nicefrac}
\usepackage{xcolor}
\usepackage{wrapfig}
\graphicspath{ {./data/} }

\usepackage{lib}

\title{On Embeddings for Numerical Features\\ in Tabular Deep Learning}

\author{%
    Yury Gorishniy\thanks{The first author: \texttt{firstnamelastname@gmail.com}} \\
    Yandex
    \And
    Ivan Rubachev \\
    HSE, Yandex
    \And
    Artem Babenko \\
    Yandex
}

\begin{document}

\maketitle

\begin{abstract}
  Recently, Transformer-like deep architectures have shown strong performance on tabular data problems. Unlike traditional models, e.g., MLP, these architectures map scalar values of numerical features to high-dimensional embeddings before mixing them in the main backbone. In this work, we argue that embeddings for numerical features are an underexplored degree of freedom in tabular DL, which allows constructing more powerful DL models and competing with gradient boosted decision trees (GBDT) on some GBDT-friendly benchmarks (that is, where GBDT outperforms conventional DL models). We start by describing two conceptually different approaches to building embedding modules: the first one is based on a piecewise linear encoding of scalar values, and the second one utilizes periodic activations. Then, we empirically demonstrate that these two approaches can lead to significant performance boosts compared to the embeddings based on conventional blocks such as linear layers and ReLU activations. Importantly, we also show that embedding numerical features is beneficial for many backbones, not only for Transformers. Specifically, after proper embeddings, simple MLP-like models can perform on par with the attention-based architectures. Overall, we highlight embeddings for numerical features as an important design aspect with good potential for further improvements in tabular DL. The source code is available at \url{https://github.com/yandex-research/tabular-dl-num-embeddings}.
\end{abstract}

\section{Introduction}


Tabular data problems are currently a final frontier for deep learning (DL) research.
While the most recent breakthroughs in NLP, vision, and speech are achieved by deep models \citep{goodfellow2016deep}, their success in the tabular domain is not convincing yet.
Despite a large number of proposed architectures for tabular DL \citep{snn, node, tabnet, autoint, grownet, tabtransformer, saint, revisiting, npt}, the performance gap between them and the ``shallow'' ensembles of decision trees, like GBDT, often remains significant \citep{revisiting, dl_is_not}.

The recent line of works \citep{revisiting, saint, npt} reduce this performance gap by successfully adapting the Transformer architecture \citep{transformer} for the tabular domain.
Compared to traditional models, like MLP or ResNet, the proposed Transformer-like architectures have a specific way to handle numerical features of the data.
Namely, they map scalar values of numerical features to high-dimensional embedding vectors, which are then mixed by the self-attention modules.
Beyond transformers, mapping numerical features to vectors was also employed in different forms in the click-through rate (CTR) prediction problems \citep{youtube,autoint,autodis}.
Nevertheless, the literature is mostly focused on developing more powerful backbones while keeping the design of embedding modules relatively simple.
In particular, the existing architectures \citep{revisiting, saint, npt, autoint, autodis} construct embeddings for numerical features using quite restrictive parametric mappings, e.g., linear functions, which can lead to suboptimal performance.
In this work, we demonstrate that the embedding step has a substantial impact on the model effectiveness, and its proper design can significantly improve tabular DL models.

Specifically, we describe two different building blocks suitable for constructing embeddings for numerical features.
The first one is a piecewise linear encoding that produces alternative initial representations for the original scalar values and is based on feature binning, a long-existing preprocessing technique \citep{discretization-review}.
The second one relies on periodic activation functions, which is inspired by their usage in implicit neural representations \citep{nerf,fourier-features,siren}, NLP \citep{transformer,dice} and CV tasks \citep{learnable-fourier}.
The first approach is simple, interpretable and non-differentiable, while the second demonstrates better results on average.
We observe that DL models equipped with our embedding schemes successfully compete with GBDT on GBDT-friendly benchmarks and achieve the new state-of-the-art on tabular DL.

As another important finding, we demonstrate that the step of embedding the numerical features is universally beneficial for different deep architectures, not only for Transformer-like ones.
In particular, we show, that after proper embeddings, simple MLP-like architectures often provide the performance comparable to the state-of-the-art attention-based models.
Overall, our work demonstrates the large impact of the embeddings of numerical features on the tabular DL performance and shows the potential of investigating more advanced embedding schemes in future research.

To sum up, our contributions are as follows:

\begin{enumerate}[nosep]
    \item We demonstrate that embedding schemes for numerical features are an underexplored research question in tabular DL. Namely, we show that more expressive embedding schemes can provide substantial performance improvements over prior models.

    \item We show that the profit from embedding numerical features is not specific for Transformer-like architectures, and proper embedding schemes benefit traditional models as well.

    \item On a number of public benchmarks, we achieve the new state-of-the-art on tabular DL.
\end{enumerate}

\section{Related work}

\textbf{Tabular deep learning.}
During several recent years, the community has proposed a large number of deep models for tabular data \citep{snn, node, tabnet, autoint, dcn, grownet, tel, tabtransformer, saint, revisiting, npt}.
However, when systematically evaluated, these models do not consistently outperform the ensembles of decision trees, such as GBDT (Gradient Boosting Decision Tree) \citep{xgboost, catboost, lightgbm}, which are typically the top-choice in various ML competitions \citep{revisiting, dl_is_not}.
Moreover, several recent works have shown that the proposed sophisticated architectures are not superior to properly tuned simple models, like MLP and ResNet \citep{revisiting, cocktails}.
In this work, unlike the prior literature, we do not aim to propose a new backbone architecture.
Instead, we focus on more accurate ways to handle numerical features, and our developments can be potentially combined with any model, including traditional MLPs and more recent Transformer-like ones.

\textbf{Transformers in tabular DL.}
Due to the tremendous success of Transformers for different domains \citep{transformer, vit}, several recent works adapt their self-attention design for tabular DL as well \citep{tabtransformer, revisiting, saint, npt}.
Compared to existing alternatives, applying self-attention modules to the numerical features of tabular data requires mapping the scalar values of these features to high-dimensional embedding vectors.
So far, the existing architectures perform this ``scalar'' $\rightarrow$ ``vector'' mapping by relatively simple computational blocks, which, in practice, can limit the model expressiveness.
For instance, the recent FT-Transformer architecture \citep{revisiting} employs only a single linear layer.
In our experiments, we demonstrate that such embedding schemes can provide suboptimal performance, and more advanced schemes often lead to substantial profit.

\textbf{CTR Prediction.}
In CTR prediction problems, objects are represented by numerical and categorical features, which makes this field highly relevant to tabular data problems.
In several works, numerical features are handled in some non-trivial way while not being the central part of the research \citep{youtube,autoint}.
Recently, however, a more advanced scheme has been proposed in \citet{autodis}.
Nevertheless, it is still based on linear layers and conventional activation functions, which we found to be suboptimal in our evaluation.

\textbf{Feature binning.}
Binning is a discretization technique that converts numerical features to categorical features.
Namely, for a given feature, its value range is split into bins (intervals), after which the original feature values are replaced with discrete descriptors (e.g. bin indices or one-hot vectors) of the corresponding bins.
We point to the work by \citet{discretization-review}, which performs an overview of some classic approaches to binning and can serve as an entry point to the relevant literature on the topic.
In our work, however, we utilize bins in a different way.
Specifically, we use their edges to construct lossless piecewise linear representations of the original scalar values.
It turns out that this simple and interpretable representations can provide substantial benefit to deep models on several tabular problems.

\textbf{Periodic activations.}
Recently, periodic activation functions have become a key component in processing coordinates-like inputs, which is required in many applications.
Examples include NLP \citep{transformer}, CV \citep{learnable-fourier}, implicit neural representations \citep{nerf,fourier-features,siren}.
In our work, we show that periodic activations can be used to construct powerful embedding modules for numerical features in tabular data problems.
Contrary to some of the aforementioned papers, where components of the multidimensional coordinates are mixed (e.g. with linear layers) before passing them to periodic functions \citep{siren,fourier-features}, we find it crucial to embed each feature separately before mixing them in the main backbone.

\section{Embeddings for numerical features}
\label{sec:method}

In this section, we describe the general framework for what we call "embeddings for numerical features" and the main building blocks used in the experimental comparison in \autoref{sec:experiments}.

\textbf{Notation.} For a given supervised learning problem on tabular data, we denote the dataset as $\left\{\left(x^j,\ y^j\right)\right\}_{j = 1}^n$ where $y^j \in \Y$ represents the object's label and $x^j{=}\left(x^{j(num)},\ x^{j(cat)}\right) \in \X$ represents the object's features (numerical and categorical).
$x_i^{j(num)}$, in turn, denotes the $i$-th numerical feature of the $j$-th object.
Depending on the context, the $j$ index can be omitted.
The dataset is split into three disjoint parts: $\overline{1,n} = J_{train} \cup J_{val} \cup J_{test}$, where the ``train'' part is used for training, the ``validation'' part is used for early stopping and hyperparameter tuning, and the ``test'' part is used for the final evaluation.

\subsection{General framework}
\label{sec:general-framework}

We formalize the notion of "embeddings for numerical features" as $z_i = f_i((x_i^{(num)}) \in \R^{d_i}$, where $f_i(x)$ is the embedding function for the $i$-th numerical feature, $z_i$ is the embedding of the $i$-th numerical feature and $d_i$ is the dimensionality of the embedding.
Importantly, the proposed framework implies that embeddings for all features are computed \textit{independently} of each other.
Note that the function $f_i$ can depend on parameters that are trained as a part of the whole model or in some other fashion (e.g. before the main optimization).
In this work, we consider only embedding schemes where the embedding functions for all features are of the same functional form.
We never share parameters of embedding functions of different features.

The subsequent use of the embeddings depends on the model backbone.
For MLP-like architectures, they are concatenated into one flat vector (see \autoref{A:mlp-with-embeddings} for illustrations).
For Transformer-based architectures, no extra step is performed and the embeddings are passed as is, so the usage is defined by the original architectures.

\subsection{Piecewise linear encoding}
\label{sec:ple}

While vanilla MLP is known to be a universal approximator \citep{universal-approximation-1989,universal-approximation-1991}, in practice, due to optimization peculiarities, it has limitations in its learning capabilities \citep{spectral-bias}.
However, the recent work by \citet{fourier-features} uncovers the case where changing the input space alleviates the above issue.
This observation motivates us to check if changing the representations of the original scalar values of numerical features can improve the learning capabilities of tabular DL models.

At this point, we try to start simple and turn to "classical" machine learning techniques. Namely, we take inspiration from the one-hot encoding algorithm that is widely and successfully used for representing discrete entities such as categorical features in tabular data problems or tokens in NLP.
We note that the one-hot representation can be seen as an opposite solution to the scalar representation in terms of the trade-off between parameter efficiency and expressivity.
To check whether the one-hot-like approach can be beneficial for tabular DL models, we design a continuous alternative to the one-hot encoding (since the vanilla one-hot encoding is barely applicable to numerical features).

Formally, for the $i$-th numerical feature, we split its value range into the disjoint set of $T^i$ intervals $B_1^i,\ \dots,\ B_T^i$, which we call \textit{bins}: $B_t^i = [b_{t - 1}^i, b_t^i)$.
The splitting algorithm is an important implementation detail that we discuss later.
From now on, we omit the feature index $i$ for simplicity.
Once the bins are determined, we define the encoding scheme as in \autoref{eq:piecewise-linear-encoding}:

\vspace{-1em}
\begin{minipage}{0.45\textwidth}
    \begin{align}
    \label{eq:piecewise-linear-encoding}
    \begin{split}
        & \PLE(x) = [e_1,\ \dots,\ e_T] \in \R^T \\
        & e_t =
            \begin{cases}
                0, & x < b_{t - 1}\ \texttt{AND}\ t > 1\\
                1, & x \ge b_t\ \texttt{AND}\ t < T\\
                \frac{x - b_{t - 1}}{b_t - b_{t - 1}}, & \text{otherwise}
            \end{cases}
    \end{split}
    \end{align}

    where \PLE\ stands for ``\textbf{p}eicewise \textbf{l}inear \textbf{e}ncoding''. We provide the visualization in \autoref{fig:ple}.
\end{minipage}
\hspace{0.05\textwidth}
\begin{minipage}{0.45\textwidth}
    \vspace{1em}
    \centering
    \includegraphics[width=0.85\linewidth]{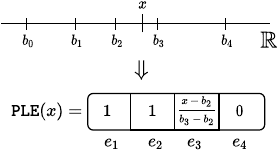}
    \captionof{figure}{The piecewise linear encoding (\PLE) in action for $T = 4$ (see \autoref{eq:piecewise-linear-encoding}).}
    \label{fig:ple}
\end{minipage}

Note that:
\begin{itemize}[nosep, leftmargin=2em]
    \item \PLE\ produces alternative initial representations for the numerical features and can be viewed as a preprocessing strategy. These representations are computed once and then used instead of the original scalar values during the main optimization.
    \item For $T = 1$, the \PLE-representation is effectively equivalent to the scalar representation.
    \item Contrary to categorical features, numerical features are ordered; we express that by setting to $1$ the components corresponding to bins with the right boundaries lower than the given feature value (this approach resembles how labels are encoded in ordinal regression problems).
    \item The cases $(x < b_0)$ and $(x \ge b_T)$ are also covered by \autoref{eq:piecewise-linear-encoding} (which leads to $(e_1 \le 0)$ and $(e_T \ge 1)$ respectively).
    \item The choice to make the representation piecewise linear is itself a subject for discussion. We analyze some alternatives in \autoref{sec:ablation}.
    \item \PLE\ can be viewed as feature preprocessing, which is additionally discussed in \autoref{sec:ple-preprocessing}.
\end{itemize}

\textbf{A note on attention-based models.}
While the described \PLE-representations can be passed to MLP-like models as is, attention-based models are inherently invariant to the order of input embeddings, so one additional step is required to add the information about feature indices to the obtained encodings.
Technically, we observe that it is enough to place one linear layer after \PLE (without sharing weights between features).
Conceptually, however, this solution has a clear semantic interpretation.
Namely, it is equivalent to allocating one trainable embedding $v_t \in \R^d$ for each bin $B_t$ and obtaining the final feature embedding by aggregating the embeddings of its bins with $e_t$ as weights, plus bias $v_0$. Formally: $f_i\left(x\right) = v_0 + \sum_{t = 1}^T e_t \cdot v_t = \Linear \left(\PLE \left(x\right)\right)$.

In the following two sections, we describe two simple algorithms for building bins suitable for \PLE.
Namely, we rely on the classic binning algorithms \citep{discretization-review} and one of the two algorithms is unsupervised, while another one utilizes labels for constructing bins.

\subsubsection{Obtaining bins from quantiles}
\label{sec:ple-quantiles}

A natural baseline way to construct the bins for \PLE\ is by splitting value ranges according to the uniformly chosen empirical quantiles of the corresponding individual feature distributions.
Formally, for the $i$-th feature: $b_t = \texttt{Q}_{\frac{t}{T}}\left(\{x_i^{j(num)}\}_{j \in J_{train}}\right)$, where \texttt{Q} is the empirical quantile function.
Trivial bins of zero size are removed. In \autoref{A:synthetic}, we demonstrate the usefulness of the proposed scheme on the synthetic GBDT-friendly dataset described in section 5.1 in \citet{revisiting}.

\subsubsection{Building target-aware bins}
\label{sec:target-aware-ple}

In fact, there are also supervised approaches that employ training labels for constructing bins \citep{discretization-review}.
Intuitively, such target-aware algorithms aim to produce bins that correspond to relatively narrow ranges of possible target values.
The supervised approach used in our work is identical in its spirit to the "C4.5 Discretization" algorithm from \citet{c4.5-disc}.
In a nutshell, for each feature, we recursively split its value range in a greedy manner using target as guidance, which is equivalent to building a decision tree (which uses for growing only this one feature and the target) and treating the regions corresponding to its leaves as the bins for \PLE\ (see the illustration in \autoref{fig:A_target-aware-ple}).
Additionally, we define $b_0^i = \min_{j \in J_{train}}x_i^j$ and $b_T^i = \max_{j \in J_{train}}x_i^j$.

\subsection{Periodic activation functions}
\label{sec:periodic}
Recall that in \autoref{sec:ple} the work by \citet{fourier-features} was used as a starting point of our motivation for developing \PLE.
Thus, we also try to adapt the original work itself for tabular data problems.
Our variation differs in two aspects.
First, we take into account the fact the embedding framework described in \autoref{sec:general-framework} forbids mixing features during the embedding process (see \autoref{A:fourier} for additional discussion).
Second, we train the pre-activation coefficients instead of keeping them fixed.
As a result, our approach is rather close to \citet{learnable-fourier} with the number of ``groups'' equal to the number of numerical features.
We formalize the described scheme in \autoref{eq:periodic},

\vspace{-1em}
\begin{equation}
\label{eq:periodic}
    f_i(x) = \texttt{Periodic}(x) = \texttt{concat}[\sin(v),\ \cos(v)], \qquad v = [2 \pi c_1 x,\ \dots,\ 2 \pi c_k x]
\end{equation}
\vspace{-1em}

where $c_i$ are trainable parameters initialized from $\mathcal{N}(0, \sigma)$.
We observe that $\sigma$ is an important hyperparameter. Both $\sigma$ and $k$ are tuned using validation sets.

\subsection{Simple differentiable layers}
In the context of Deep Learning, embedding numerical features with conventional differentiable layers (e.g. linear layers, ReLU activation, etc.) is a natural approach.
In fact, this technique is already used on its own in the recently proposed attention-based architectures \citep{revisiting, npt, saint} and in some models for CTR prediction problems \citep{autoint,autodis}.
However, we also note that such conventional modules can be used on top of the components described in \autoref{sec:ple} and \autoref{sec:periodic}.
In \autoref{sec:experiments}, we find that such combinations often lead to better results.

\section{Experiments}
\label{sec:experiments}

In this section, we empirically evaluate the techniques discussed in \autoref{sec:method} and compare them with Gradient Boosted Decision Trees to check the status quo of the ``DL vs GBDT'' competition.

\subsection{Datasets}

\begin{table}[h]
    \setlength\tabcolsep{2.2pt}
    \centering
    \caption{Dataset properties. ``RMSE'' denotes root-mean-square error, ``Acc.'' denotes accuracy.}
    \label{tab:datasets}
    \vspace{1em}
    \scalebox{0.95}{{\footnotesize \begin{tabular}{lcccccccccccc}
\toprule
{} & GE & CH & CA & HO & AD & OT & HI & FB & SA & CO & MI \\
\midrule
\text{\#objects} & 9873 & 10000 & 20640 & 22784 & 48842 & 61878 & 98049 & 197080 & 200000 & 581012 & 1200192 \\
\#num. features & 32 & 10 & 8 & 16 & 6 & 93 & 28 & 50 & 200 & 54 & 136 \\
\#cat. features & 0 & 1 & 0 & 0 & 8 & 0 & 0 & 1 & 0 & 0 & 0 \\
metric & Acc. & Acc. & RMSE & RMSE & Acc. & Acc. & Acc. & RMSE & Acc. & Acc. & RMSE \\
\#classes & 5 & 2 & -- & -- & 2 & 9 & 2 & -- & 2 & 7 & -- \\
majority class & 29\% & 79\% & -- & -- & 76\% & 26\% & 52\% & -- & 89\% & 48\% & -- \\
\bottomrule
\end{tabular}}}
\end{table}

We use eleven public datasets mostly from the previous works on tabular DL and Kaggle competitions.
Importantly, we focus on the middle and large scale tasks, and our benchmark is biased towards GBDT-friendly problems, since, as of now, closing the gap with GBDT models on such tasks is one of the main challenges for tabular DL.
The main dataset properties are summarized in \autoref{tab:datasets} and the used sources and additional details are provided in \autoref{A:datasets}.

\subsection{Implementation details}
\label{sec:implementation-details}

We mostly follow \citet{revisiting} in terms of the hyperparameter tuning, training and evaluation protocols.
Nevertheless, for completeness, we list all the details in \autoref{A:implementation-details}.
In the next paragraph, we describe the implementation details specific to embeddings for numerical features.

\textbf{Embeddings for numerical features.}
If linear layers are used, we tune their output dimensions.
The \PLE\ hyperparameters are the same for all features.
For quantile-based \PLE, we tune the number of quantiles.
For target-aware \PLE, we tune the following parameters for decision trees: the maximum number of leaves, the minimum number of items per leaf, and the minimum information gain required for making a split when growing the tree.
For the \Periodic\ module (see \autoref{eq:periodic}), we tune $\sigma$ and $k$ (these hyperparameters are the same for all features).

\subsection{Model names}
\label{sec:names}

\begin{minipage}[t]{0.425\linewidth}
In the experiments, we consider different combinations of backbones and embeddings.
For convenience, we use the ``Backbone-Embedding'' pattern to name the models, where ``Backbone'' denotes the backbone (e.g. MLP, ResNet, Transformer) and ``Embedding'' denotes the embedding type.
See \autoref{tab:names} for all considered embedding modules.
Note that:
\begin{itemize}[nosep,leftmargin=*]
    \item \Periodic\ is defined in \autoref{eq:periodic}.
    \item $\PLE_{\texttt{q}}$ denotes the quantile-based \PLE. $\PLE_{\texttt{t}}$ denotes the target-aware \PLE.
    \item $\Linear_{-}$ denotes bias-free linear layer. \texttt{LReLU} denotes leaky ReLU. \texttt{AutoDis} was proposed in \citet{autodis}
    \item ``Transformer-L'' is equivalent to FT-Transformer \citep{revisiting}.
\end{itemize}
\end{minipage}
\hspace{0.025\linewidth}
\begin{minipage}[t]{0.55\linewidth}
    \centering
    \captionof{table}{Embedding names. See \autoref{sec:names}}
    \scalebox{0.88}{{\footnotesize
    \begin{tabular}{r|r}
         Name & Embedding function ($f_i$) \\
         \midrule
         \texttt{L} & \Linear \\
         \texttt{LR} & $\ReLU \circ \Linear$ \\
         \texttt{LRLR} & $\ReLU \circ \Linear \circ \ReLU \circ \Linear$ \\
         \texttt{Q} & $\PLE_{\texttt{q}}$ \\
         \texttt{Q-L} & $\Linear \circ \PLE_{\texttt{q}}$ \\
         \texttt{Q-LR} & $\ReLU \circ \Linear \circ \PLE_{\texttt{q}}$ \\
         \texttt{Q-LRLR} & $\ReLU \circ \Linear \circ \ReLU \circ \Linear \circ \PLE_{\texttt{q}}$ \\
         \texttt{T} & $\PLE_{\texttt{t}}$ \\
         \texttt{T-L} & $\Linear \circ \PLE_{\texttt{t}}$ \\
         \texttt{T-LR} & $\ReLU \circ \Linear \circ \PLE_{\texttt{t}}$ \\
         \texttt{T-LRLR} & $\ReLU \circ \Linear \circ \ReLU \circ \Linear \circ \PLE_{\texttt{t}}$ \\
         \texttt{P} & $\Periodic$ \\
         \texttt{PL} & $\Linear \circ \Periodic$ \\
         \texttt{PLR} & $\ReLU \circ \Linear \circ \Periodic$ \\
         \texttt{PLRLR} & $\ReLU \circ \Linear \circ \ReLU \circ \Linear \circ \Periodic$ \\
         \texttt{AutoDis} & $\Linear \circ \texttt{SoftMax} \circ \Linear_{-} \circ \texttt{LReLU} \circ \Linear_{-}$
    \end{tabular}
    }}
    \label{tab:names}
\end{minipage}

\subsection{Simple differentiable embedding modules}
\label{sec:mlp}

\begin{table}[h]
    \setlength\tabcolsep{2.2pt}
    \centering
    \caption{Results for MLP equipped with simple embedding modules (see \autoref{sec:names}).
    The metric values averaged over 15 random seeds are reported.
    The standard deviations are provided in \autoref{A:stds}.
    We consider one result to be better than another if its mean score is better and its standard deviation is less than the difference.
    For each dataset, top results are in \textbf{bold}.
    Notation: \textdownarrow\ corresponds to RMSE, \textuparrow\ corresponds to accuracy}
    \label{tab:simple-embeddings}
    \vspace{1em}
    {\footnotesize \begin{tabular}{lccccccccccc}
\toprule
{} & GE \textuparrow & CH \textuparrow & CA \textdownarrow & HO \textdownarrow & AD \textuparrow & OT \textuparrow & HI \textuparrow & FB \textdownarrow & SA \textuparrow & CO \textuparrow & MI \textdownarrow \\
\midrule
MLP             & $\mathbf{0.632}$ & $0.856$ & $0.495$ & $3.204$ & $0.854$ & $0.818$ & $0.720$ & $5.686$ & $0.912$ & $\mathbf{0.964}$ & $0.747$ \\
MLP-L           & $\mathbf{0.639}$ & $\mathbf{0.861}$ & $0.475$ & $3.123$ & $\mathbf{0.856}$ & $\mathbf{0.820}$ & $0.723$ & $5.684$ & $0.916$ & $0.963$ & $0.748$ \\
MLP-LR          & $\mathbf{0.642}$ & $\mathbf{0.860}$ & $\mathbf{0.471}$ & $\mathbf{3.084}$ & $\mathbf{0.857}$ & $\mathbf{0.819}$ & $\mathbf{0.726}$ & $\mathbf{5.625}$ & $\mathbf{0.923}$ & $0.963$ & $\mathbf{0.746}$ \\
\bottomrule
\end{tabular}}
\end{table}

We start by evaluating embedding modules consisting of ``conventional'' differentiable layers (linear layers, ReLU activations, etc.).
The results are summarized in \autoref{tab:simple-embeddings}.
\\\textbf{The main takeaways}:
\begin{itemize}[nosep,leftmargin=2em]
    \item first and foremost, the results indicate that MLP can benefit from embedding modules. Thus, we conclude that this backbone is worth attention when it comes to evaluating embedding modules.
    \item the simple \texttt{LR} module leads to modest, but consistent improvements when applied to MLP.
\end{itemize}

Interestingly, the ``redundant'' MLP-L configuration also tends to outperform the vanilla MLP.
Although the improvements are not dramatic, the special property of this architecture is that the linear embedding module can be fused together with the first linear layer of MLP after training, which completely removes the overhead.
As for \texttt{LRLR} and AutoDis, we observe that these heavy modules do not justify the extra costs (see the results in \autoref{A:stds}).

\subsection{Piecewise linear encoding}

In this section, we evaluate the encoding scheme described in \autoref{sec:ple}.
The results are summarized in \autoref{tab:ple}.
\\\textbf{The main takeaways}:
\begin{itemize}[nosep, leftmargin=2em]
    \item The piecewise linear encoding is often beneficial for both types of architectures (MLP and Transformer) and the profit can be significant (for example, see the CA and AD datasets).
    \item Adding differentiable components on top of the \PLE\ can improve the performance. Though, the most expensive modifications such as \texttt{Q-LRLR} and \texttt{T-LRLR} are not worth it (see \autoref{A:stds}).
\end{itemize}

\newpage
Note that the benchmark is biased towards GBDT-friendly problems, so the typical superiority of tree-based bins over quantile-based bins, which can be observed in \autoref{tab:ple}, may not generalize to more DL-friendly datasets.
Thus, we do not make any general claims about the relative advantages of the two schemes here.

\begin{table}[h]
    \setlength\tabcolsep{2.2pt}
    \centering
    \caption{
    Results for MLP and Transformer with embedding modules based on the piecewise linear encoding (\autoref{sec:ple}).
    Notation follows \autoref{tab:simple-embeddings} and \autoref{tab:names}.
    The best results are defined separately for the MLP and Transformer backbones.
    }
    \label{tab:ple}
    \vspace{1em}
    \scalebox{0.95}{{\footnotesize \begin{tabular}{lccccccccccc}
\toprule
{} & GE \textuparrow & CH \textuparrow & CA \textdownarrow & HO \textdownarrow & AD \textuparrow & OT \textuparrow & HI \textuparrow & FB \textdownarrow & SA \textuparrow & CO \textuparrow & MI \textdownarrow \\
\midrule
MLP             & $0.632$ & $0.856$ & $0.495$ & $3.204$ & $0.854$ & $0.818$ & $0.720$ & $5.686$ & $0.912$ & $0.964$ & $\mathbf{0.747}$ \\
MLP-Q           & $\mathbf{0.653}$ & $0.854$ & $0.464$ & $\mathbf{3.163}$ & $0.859$ & $0.816$ & $0.721$ & $5.766$ & $0.922$ & $0.968$ & $0.750$ \\
MLP-T           & $\mathbf{0.647}$ & $\mathbf{0.861}$ & $0.447$ & $\mathbf{3.149}$ & $0.864$ & $\mathbf{0.821}$ & $0.720$ & $5.577$ & $0.923$ & $0.967$ & $0.749$ \\
MLP-Q-LR        & $\mathbf{0.646}$ & $0.857$ & $0.455$ & $\mathbf{3.184}$ & $0.863$ & $0.811$ & $0.720$ & $\mathbf{5.394}$ & $0.923$ & $\mathbf{0.969}$ & $\mathbf{0.747}$ \\
MLP-T-LR        & $0.640$ & $\mathbf{0.861}$ & $\mathbf{0.439}$ & $3.207$ & $\mathbf{0.868}$ & $0.818$ & $\mathbf{0.724}$ & $\mathbf{5.508}$ & $\mathbf{0.924}$ & $\mathbf{0.968}$ & $0.747$ \\
\midrule
Transformer-L   & $0.632$ & $0.860$ & $0.465$ & $3.239$ & $0.858$ & $\mathbf{0.817}$ & $0.725$ & $\mathbf{5.602}$ & $0.924$ & $0.971$ & $\mathbf{0.746}$ \\
Transformer-Q-L & $\mathbf{0.659}$ & $0.856$ & $0.451$ & $3.319$ & $0.867$ & $0.812$ & $\mathbf{0.729}$ & $5.741$ & $\mathbf{0.924}$ & $\mathbf{0.973}$ & $0.747$ \\
Transformer-T-L & $\mathbf{0.663}$ & $\mathbf{0.861}$ & $0.454$ & $\mathbf{3.197}$ & $\mathbf{0.871}$ & $\mathbf{0.817}$ & $0.726$ & $5.803$ & $\mathbf{0.924}$ & $\mathbf{0.974}$ & $0.747$ \\
Transformer-Q-LR & $\mathbf{0.659}$ & $0.857$ & $0.448$ & $3.270$ & $0.867$ & $0.812$ & $0.723$ & $5.683$ & $0.923$ & $0.972$ & $0.748$ \\
Transformer-T-LR & $\mathbf{0.665}$ & $0.860$ & $\mathbf{0.442}$ & $\mathbf{3.219}$ & $0.870$ & $\mathbf{0.818}$ & $\mathbf{0.729}$ & $5.699$ & $\mathbf{0.924}$ & $0.973$ & $0.747$ \\
\bottomrule
\end{tabular}}}
\end{table}

\subsection{Periodic activation functions}

\begin{table}[h]
    \setlength\tabcolsep{2.2pt}
    \centering
    \caption{
    Results for MLP and Transformer with embedding modules based on periodic activations (\autoref{sec:periodic}).
    Notation follows \autoref{tab:simple-embeddings} and \autoref{tab:names}.
    The best results are defined separately for the MLP and Transformer backbones.
    }
    \vspace{1em}
    \label{tab:periodic}
    \scalebox{0.95}{{\footnotesize \begin{tabular}{lccccccccccc}
\toprule
{} & GE \textuparrow & CH \textuparrow & CA \textdownarrow & HO \textdownarrow & AD \textuparrow & OT \textuparrow & HI \textuparrow & FB \textdownarrow & SA \textuparrow & CO \textuparrow & MI \textdownarrow \\
\midrule
MLP             & $0.632$ & $0.856$ & $0.495$ & $3.204$ & $0.854$ & $\mathbf{0.818}$ & $0.720$ & $5.686$ & $0.912$ & $0.964$ & $0.747$ \\
MLP-P           & $0.631$ & $\mathbf{0.860}$ & $0.489$ & $3.129$ & $0.869$ & $0.807$ & $0.723$ & $5.845$ & $0.923$ & $0.968$ & $0.747$ \\
MLP-PL          & $0.641$ & $\mathbf{0.859}$ & $\mathbf{0.467}$ & $3.113$ & $0.868$ & $\mathbf{0.819}$ & $0.727$ & $\mathbf{5.530}$ & $0.924$ & $\mathbf{0.969}$ & $0.746$ \\
MLP-PLR         & $\mathbf{0.674}$ & $\mathbf{0.857}$ & $\mathbf{0.467}$ & $\mathbf{3.050}$ & $\mathbf{0.870}$ & $\mathbf{0.819}$ & $\mathbf{0.728}$ & $\mathbf{5.525}$ & $\mathbf{0.924}$ & $\mathbf{0.970}$ & $\mathbf{0.746}$ \\
\midrule
Transformer-L   & $\mathbf{0.632}$ & $0.860$ & $\mathbf{0.465}$ & $3.239$ & $0.858$ & $\mathbf{0.817}$ & $0.725$ & $\mathbf{5.602}$ & $\mathbf{0.924}$ & $\mathbf{0.971}$ & $\mathbf{0.746}$ \\
Transformer-PLR & $\mathbf{0.646}$ & $\mathbf{0.863}$ & $\mathbf{0.464}$ & $\mathbf{3.162}$ & $\mathbf{0.870}$ & $0.814$ & $\mathbf{0.730}$ & $5.760$ & $\mathbf{0.924}$ & $\mathbf{0.972}$ & $\mathbf{0.746}$ \\
\bottomrule
\end{tabular}}}
\end{table}

In this section, we evaluate embedding modules based on periodic activation functions as described in \autoref{sec:periodic}.
The results are reported in \autoref{tab:periodic}.
\\\textbf{The main takeaway}: on average, MLP-P is superior to the vanilla MLP. However, adding a differentiable component on top of the \Periodic\ module should be the default strategy (which is in line with \citet{learnable-fourier}).
Indeed, MLP-PLR and MLP-PL provide meaningful improvements over MLP-P (e.g. see GE, CA, HO) and even ``fix'' MLP-P where it is inferior to MLP (OT, FB).

Although MLP-PLR is usually superior to MLP-PL, we note that in the latter case the last linear layer of the embedding module is ``redundant'' in terms of expressivity and can be fused with the first linear layer of the backbone after training, which, in theory, can lead to a more lightweight model.
Finally, we observe that MLP-PLRLR and MLP-PLR do not differ significantly enough to justify the extra cost of the \texttt{PLRLR} module (see \autoref{A:stds}).

\subsection{Comparing DL models and GBDT}
\label{sec:dl-gbdt}

In this section, we perform a big comparison of different approaches to identify the best embedding modules and backbones, as well as to check if embeddings for numerical features allow DL models to compete with GBDT on more tasks than before.
Importantly, we compare \textit{ensembles} of DL models against \textit{ensembles} of GBDT, since Gradient Boosting is essentially an ensembling technique, so such comparison will be fairer.
Note that we focus only on the best metric values without taking efficiency into account, so we only check if DL models are conceptually ready to compete with GBDT.

We consider three backbones: MLP, ResNet, and Transformer, since they are reported to be representative of what baseline DL backbones are currently capable of \citep{revisiting,cocktails,saint,npt}.
Note that we do not include the attention-based models that also apply attention on the level of \textit{objects} \citep{saint, npt, hopfield}, since this non-parametric component is orthogonal to the central topic of our work.
The results are summarized in \autoref{tab:dl-gbdt}.

\begin{table}[h]
    \setlength\tabcolsep{2.2pt}
    \centering
    \caption{Results for ensembles of GBDT, the baseline DL models and their modifications using different types of embeddings for numerical features. Notation follows \autoref{tab:simple-embeddings} and \autoref{tab:names}. Due to the limited precision, some \textit{different} values are represented with the same figures.}
    \label{tab:dl-gbdt}
    \vspace{1em}
    {\scriptsize \begin{tabular}{lccccccccccc|r}
\toprule
{} & GE \textuparrow & CH \textuparrow & CA \textdownarrow & HO \textdownarrow & AD \textuparrow & OT \textuparrow & HI \textuparrow & FB \textdownarrow & SA \textuparrow & CO \textuparrow & MI \textdownarrow & Avg. Rank
\\
\midrule
CatBoost        & $0.692$ & $0.861$ & $0.430$ & $3.093$ & $0.873$ & $0.825$ & $0.727$ & $5.226$ & $0.924$ & $0.967$ & $\mathbf{0.741}$ & $3.6 \pm 2.9$
\\
XGBoost         & $0.683$ & $0.859$ & $0.434$ & $3.152$ & $\mathbf{0.875}$ & $0.827$ & $0.726$ & $5.338$ & $0.919$ & $0.969$ & $0.742$ & $4.6 \pm 2.7$
\\
\midrule
MLP             & $0.665$ & $0.856$ & $0.486$ & $3.109$ & $0.856$ & $0.822$ & $0.727$ & $5.616$ & $0.913$ & $0.968$ & $0.746$ & $8.5 \pm 2.6$
\\
MLP-LR          & $0.679$ & $0.861$ & $0.463$ & $3.012$ & $0.859$ & $0.826$ & $0.731$ & $5.477$ & $0.924$ & $0.972$ & $0.744$ & $5.5 \pm 2.7$
\\
MLP-Q-LR        & $0.682$ & $0.859$ & $0.433$ & $3.080$ & $0.867$ & $0.818$ & $0.724$ & $\mathbf{5.144}$ & $0.924$ & $0.974$ & $0.745$ & $5.1 \pm 1.9$
\\
MLP-T-LR        & $0.673$ & $0.861$ & $0.435$ & $3.099$ & $0.870$ & $0.821$ & $0.727$ & $5.409$ & $0.924$ & $0.973$ & $0.746$ & $5.1 \pm 1.7$
\\
MLP-PLR         & $\mathbf{0.700}$ & $0.858$ & $0.453$ & $\mathbf{2.975}$ & $0.874$ & $\mathbf{0.830}$ & $\mathbf{0.734}$ & $5.388$ & $\mathbf{0.924}$ & $0.975$ & $0.743$ & $3.0 \pm 2.4$
\\
\midrule
ResNet          & $0.690$ & $0.861$ & $0.483$ & $3.081$ & $0.856$ & $0.821$ & $0.734$ & $5.482$ & $0.918$ & $0.968$ & $0.745$ & $6.7 \pm 3.3$
\\
ResNet-LR       & $0.672$ & $0.862$ & $0.450$ & $2.992$ & $0.859$ & $0.822$ & $0.733$ & $5.415$ & $0.923$ & $0.971$ & $0.743$ & $5.6 \pm 2.7$
\\
ResNet-Q-LR     & $0.674$ & $0.859$ & $0.427$ & $3.066$ & $0.868$ & $0.815$ & $0.729$ & $5.309$ & $0.923$ & $0.976$ & $0.746$ & $4.7 \pm 2.0$
\\
ResNet-T-LR     & $0.683$ & $0.862$ & $\mathbf{0.425}$ & $3.030$ & $0.872$ & $0.822$ & $0.731$ & $5.471$ & $0.923$ & $0.975$ & $0.744$ & $4.1 \pm 1.9$
\\
ResNet-PLR      & $0.691$ & $0.861$ & $0.443$ & $3.040$ & $\mathbf{0.874}$ & $0.825$ & $0.734$ & $5.400$ & $0.924$ & $0.975$ & $0.743$ & $3.2 \pm 1.3$
\\
\midrule
Transformer-L   & $0.668$ & $0.861$ & $0.455$ & $3.188$ & $0.860$ & $0.824$ & $0.727$ & $5.434$ & $0.924$ & $0.973$ & $0.743$ & $5.9 \pm 2.2$
\\
Transformer-LR  & $0.666$ & $0.861$ & $0.446$ & $3.193$ & $0.861$ & $0.824$ & $0.733$ & $5.430$ & $0.924$ & $0.973$ & $0.743$ & $5.2 \pm 2.2$
\\
Transformer-Q-LR & $0.690$ & $0.857$ & $\mathbf{0.425}$ & $3.143$ & $0.868$ & $0.818$ & $0.726$ & $5.471$ & $\mathbf{0.924}$ & $0.975$ & $0.744$ & $4.4 \pm 2.2$
\\
Transformer-T-LR & $0.686$ & $0.862$ & $\mathbf{0.423}$ & $3.149$ & $0.871$ & $0.823$ & $0.733$ & $5.515$ & $0.924$ & $\mathbf{0.976}$ & $0.744$ & $3.7 \pm 2.2$
\\
Transformer-PLR & $0.686$ & $\mathbf{0.864}$ & $0.449$ & $3.091$ & $0.873$ & $0.823$ & $0.734$ & $5.581$ & $\mathbf{0.924}$ & $0.975$ & $0.743$ & $3.9 \pm 2.5$
\\
\bottomrule
\end{tabular}}
\end{table}

\textbf{The main takeaways for DL models}:
\begin{itemize}[leftmargin=2em]
    \item For most datasets, embeddings for numerical features can provide noticeable improvements for three different backbones. Although the average rank is not a good metric for making subtle conclusions, we highlight the impressive difference in average ranks between the MLP and MLP-PLR models.
    \item The simplest \texttt{LR} embedding is a good baseline solution: although the performance gains are not dramatic, its main advantage is consistency (e.g. see MLP vs MLP-LR).
    \item The \texttt{PLR} module provides the best average performance. Empirically, we observe $\sigma$ (see \autoref{eq:periodic}) to be an important hyperparameter that should be tuned.
    \item Piecewise linear encoding (\PLE) allows building well performing embeddings (e.g. \text{T-LR}, \text{Q-LR}). In addition to that, \PLE\ itself is worth attention because of its simplicity, interpretability and efficiency (no computationally expensive periodic functions).
    \item Importantly, after the MLP-like architectures are coupled with embeddings for numerical features, they perform on par with the Transformer-based models.
\end{itemize}

\textbf{The main takeaway for the ``DL vs GBDT'' competition}: embeddings for numerical features is a significant design aspect that has a great potential for improving DL models and closing the gap with GBDT on GBDT-friendly tasks.
Let us illustrate this claim with several observations:
\begin{itemize}[leftmargin=2em]
    \item The benchmark is initially biased to GBDT-friendly problems, which can be observed by comparing GBDT solutions with the vanilla DL models (MLP, ResNet, Transformer-L).
    \item However, for the vast majority of the “backbone \& dataset” pairs, proper embeddings are the only thing needed to close the gap with GBDT. Exceptions (rather formal) include the MI dataset and the following pairs: \mbox{“ResNet \& GE”}, \mbox{“Transformer \& FB”}, \mbox{“Transformer \& GE”}, \mbox{“Transformer \& OT”}.
    \item Additionally, to the best of our knowledge, it is the first time when DL models perform on par with GBDT on the well-known California Housing and Adult datasets.
\end{itemize}

That said, compared to GBDT models, efficiency can still be an issue for the considered DL architectures.
In any case, the trade-off completely depends on the specific use case and requirements.

\section{Analysis}

\subsection{Comparing model sizes}
\label{sec:model-sizes}

To quantify the effect of embeddings for numerical features on model sizes, we report the parameter counts in \autoref{tab:model-sizes}.
Overall, introducing embeddings for numerical features can cause non-negligible overhead in terms of model size.
Importantly, the overhead in terms of size does not translate to the same overhead in terms of training times and throughput.
For example, the almost $2000$-fold increase in the parameter count for MLP-LR on the CH dataset results in only $1.5$-fold increase in training times.
Finally, in practice, we observe that coupling MLP and ResNet with embedding modules leads to architectures that are still faster than Transformer-based models.

\begin{table}[h]
    \setlength\tabcolsep{2.2pt}
    \centering
    \caption{Parameter counts for MLP with different embedding modules. All the models are tuned and the corresponding backbones are not identical in their sizes, so we take into account the fact that different approaches require a different number of parameters to realize their full potential.}
    \label{tab:model-sizes}
    \vspace{1em}
    \scalebox{0.93}{{\footnotesize \begin{tabular}{lccccccccccc}
\toprule
{} & GE & CH & CA & HO & AD & OT & HI & FB & SA & CO & MI \\
\midrule
MLP & $2.0$M & $1.5$K & $43.5$K & $3.6$M & $5.3$M & $479.9$K & $25.8$K & $937.3$K & $5.8$M & $3.2$M & $276.5$K \\
MLP-LR & $\times 2.52$ & $\times 1931.03$ & $\times 25.05$ & $\times 1.28$ & $\times 0.35$ & $\times 12.53$ & $\times 68.16$ & $\times 4.76$ & $\times 1.58$ & $\times 0.72$ & $\times 15.79$ \\
MLP-T & $\times 1.58$ & $\times 14.13$ & $\times 7.97$ & $\times 0.43$ & $\times 0.04$ & $\times 2.27$ & $\times 5.85$ & $\times 0.47$ & $\times 0.59$ & $\times 0.74$ & $\times 3.85$ \\
MLP-T-LR & $\times 1.61$ & $\times 463.55$ & $\times 6.80$ & $\times 0.23$ & $\times 0.16$ & $\times 2.52$ & $\times 113.22$ & $\times 3.43$ & $\times 0.41$ & $\times 0.35$ & $\times 8.47$ \\
MLP-PLR & $\times 1.73$ & $\times 250.24$ & $\times 12.94$ & $\times 1.07$ & $\times 0.66$ & $\times 8.05$ & $\times 110.57$ & $\times 4.93$ & $\times 0.64$ & $\times 0.44$ & $\times 9.57$ \\
\bottomrule
\end{tabular}
}}
\end{table}

\subsection{Ablation study}
\label{sec:ablation}

\begin{table}[h]
    \setlength\tabcolsep{2.2pt}
    \centering
    \caption{Comparing piecewise linear encoding (\PLE) with the two variations described in \autoref{sec:ablation}. Notation follows \autoref{tab:simple-embeddings} and \autoref{tab:names}.}
    \label{tab:ablation}
    \vspace{1em}
    {\footnotesize \begin{tabular}{lcccccccc}
\toprule
{} & GE \textuparrow & CH \textuparrow & CA \textdownarrow & HO \textdownarrow & AD \textuparrow & OT \textuparrow & HI \textuparrow & FB \textdownarrow \\
\midrule
MLP-Q (piecewise linear) & $\mathbf{0.653}$ & $\mathbf{0.854}$ & $0.464$ & $\mathbf{3.163}$ & $\mathbf{0.859}$ & $\mathbf{0.816}$ & $\mathbf{0.721}$ & $5.766$ \\
MLP-Q (binary)  & $\mathbf{0.652}$ & $0.815$ & $0.462$ & $3.200$ & $\mathbf{0.860}$ & $0.810$ & $\mathbf{0.720}$ & $5.748$ \\
MLP-Q (one-blob) & $0.613$ & $0.851$ & $\mathbf{0.461}$ & $\mathbf{3.187}$ & $0.857$ & $0.808$ & $0.719$ & $\mathbf{5.645}$ \\
\midrule
MLP-T (piecewise linear) & $\mathbf{0.647}$ & $\mathbf{0.861}$ & $\mathbf{0.447}$ & $\mathbf{3.149}$ & $0.864$ & $\mathbf{0.821}$ & $0.720$ & $5.577$ \\
MLP-T (binary)  & $0.639$ & $0.855$ & $0.464$ & $\mathbf{3.163}$ & $\mathbf{0.869}$ & $0.813$ & $0.718$ & $5.572$ \\
MLP-T (one-blob) & $0.622$ & $0.858$ & $0.464$ & $\mathbf{3.158}$ & $\mathbf{0.870}$ & $0.809$ & $\mathbf{0.724}$ & $\mathbf{5.475}$ \\
\bottomrule
\end{tabular}}
\end{table}

In this section, we compare two alternative binning-based encoding schemes with \PLE\ (see \autoref{sec:ple}).
The first one ("thermometer" \cite{thermometer}) sets the value $1$ instead of the piecewise linear term (see \autoref{eq:piecewise-linear-encoding}).
The second one is a generalized version of the one-blob encoding \citep{one-blob} (see \autoref{A:one-blob} for details).
The tuning and evaluation protocols are the same as in \autoref{sec:implementation-details}.
The results in table \autoref{tab:ablation} indicate that making the binning-based encoding piecewise linear is a good default strategy.

\subsection{Piecewise linear encoding as a feature preprocessing technique}
\label{sec:ple-preprocessing}

\begin{table}[h]
    \setlength\tabcolsep{2.2pt}
    \centering    \caption{Results for MLP and MLP with \PLE\ for different types of data preprocessing. Solutions using \PLE\ are significantly less sensitive to data preprocessing. Notation follows \autoref{tab:simple-embeddings} and \autoref{tab:names}.}
    \label{tab:A_preprocessing}
    \vspace{1em}
    {\footnotesize \begin{tabular}{lcccccccccc}
\toprule
{} & GE \textuparrow & CH \textuparrow & CA \textdownarrow & HO \textdownarrow & AD \textuparrow & HI \textuparrow & FB \textdownarrow & SA \textuparrow & CO \textuparrow & MI \textdownarrow \\
\midrule
MLP (none)      & $0.565$ & $0.796$ & $1.118$ & $5.328$ & $0.808$ & $0.707$ & $13.125$ & $0.911$ & $0.948$ & $0.844$ \\
MLP (standard)  & $0.629$ & $0.855$ & $0.509$ & $3.303$ & $0.855$ & $0.721$ & $5.919$ & $0.912$ & $0.963$ & $0.754$ \\
MLP (quantile)  & $0.632$ & $0.856$ & $0.495$ & $3.204$ & $0.854$ & $0.720$ & $5.686$ & $0.912$ & $0.964$ & $0.747$ \\
\midrule
MLP-Q (none)    & $0.654$ & $0.851$ & $0.463$ & $3.162$ & $0.860$ & $0.721$ & $5.889$ & $0.922$ & $0.968$ & $0.754$ \\
MLP-Q (quantile) & $0.653$ & $0.854$ & $0.464$ & $3.163$ & $0.859$ & $0.721$ & $5.766$ & $0.922$ & $0.968$ & $0.750$ \\
\midrule
MLP-T (none)    & $0.644$ & $0.860$ & $0.447$ & $3.175$ & $0.865$ & $0.721$ & $5.598$ & $0.923$ & $0.968$ & $0.749$ \\
MLP-T (quantile) & $0.647$ & $0.861$ & $0.447$ & $3.149$ & $0.864$ & $0.720$ & $5.577$ & $0.923$ & $0.967$ & $0.749$ \\
\bottomrule
\end{tabular}}
\end{table}

It is known that data preprocessing, such as standardization or quantile transformation, is often crucial for DL models for achieving competitive performance.
Moreover, the performance can significantly vary between different types of preprocessing.
At the same time, \PLE-representations contain only values from $[0,\ 1]$ and they are invariant to shifting and scaling, which makes \PLE\ itself a general feature preprocessing technique potentially suitable for DL models without the need to use traditional preprocessing first.

To illustrate that, for datasets where the quantile transformation was used in \autoref{sec:experiments}, we reevaluate the tuned configurations of MLP, MLP-Q, and MLP-T with different preprocessing policies and report the results in \autoref{tab:A_preprocessing} (note that standardization is equivalent to no preprocessing for models with \PLE).
\\First, the vanilla MLP often becomes unusable without preprocessing.
Second, for the vanilla MLP, it can be important to choose one specific type of preprocessing (CA, HO, FB, MI), which is less pronounced for MLP-Q and not the case for MLP-T (though, this specific observation can be the property of the benchmarks, not of MLP-T).
Overall, the results indicate that models using \PLE\ are less sensitive to the initial preprocessing compared to the vanilla MLP.
This is an additional benefit of \PLE-representations for practitioners since the aspect of preprocessing becomes less critical with \PLE.

\subsection{The ``feature engineering'' perspective}

\begin{table}[h]
    \setlength\tabcolsep{2.2pt}
    \centering    \caption{The comparison of the effects of \Periodic-based modules for XGBoost and MLP}
    \label{tab:feature-engineering}
    \vspace{1em}
    {\footnotesize
    \begin{tabular}{lccc}
    \toprule
    {} & CA \textdownarrow & HO \textdownarrow & HI \textuparrow \\
    \midrule
    XGBoost & 0.436 & 3.160 & 0.724 \\
    XGBoost with \Periodic & 0.441 & 3.184 & 0.724 \\
    \midrule
    MLP & 0.495 & 3.204 & 0.720 \\
    MLP-PL & 0.467 & 3.113 & 0.727 \\
    \bottomrule
    \end{tabular}
    }
\end{table}

At first sight, feature embeddings may resemble feature engineering and should be suitable for all kinds of models.
However, the proposed embedding schemes are motivated by DL-specific aspects of training (see the motivational parts of \autoref{sec:ple} and \autoref{sec:periodic}).
While our methods are likely to transfer well to models with similar training properties (e.g. to linear models since those are a special case of deep models), it is not the case in general.
To illustrate that, we try adopting the \Periodic\ module for XGBoost by fixing the random coefficients from \autoref{eq:periodic}.
We also keep the original features instead of dropping them.
The tuning and evaluation protocols are the same as in \autoref{sec:implementation-details}.
The results in \autoref{tab:feature-engineering} show that this technique, while being useful for DL models, does not provide any benefits for XGBoost.

\section{Conclusion {\&} Future work}
In this work, we have demonstrated that embeddings for numerical features are an important design aspect of tabular DL architectures.
Namely, it allows existing DL backbones to achieve noticeably better results and significantly reduce the gap with Gradient Boosted Decision Trees.
We have described two approaches illustrating this phenomenon, one using the piecewise linear encoding of original scalar values, and another using periodic functions.
We have also shown that traditional MLP-like models coupled with embeddings can perform on par with attention-based models.

Nevertheless, we have only scratched the surface of the new direction. For example, it is still to be explained how exactly the discussed embedding modules help optimization on the fundamental level. Additionally, we have considered only schemes where the same functional transformation was applied to all features, which may be a suboptimal choice.

\medskip
\bibliographystyle{abbrvnat}
\bibliography{references}

\section*{Checklist}

\begin{enumerate}

\item For all authors...
\begin{enumerate}
  \item Do the main claims made in the abstract and introduction accurately reflect the paper's contributions and scope?
    \answerYes{}
  \item Did you describe the limitations of your work?
    \answerYes{See the analysis in \autoref{sec:model-sizes}.}
  \item Did you discuss any potential negative societal impacts of your work?
    \answerNA{The work focuses on a generic aspect of deep learning models.}
  \item Have you read the ethics review guidelines and ensured that your paper conforms to them?
    \answerYes{}
\end{enumerate}

\item If you are including theoretical results...
\begin{enumerate}
  \item Did you state the full set of assumptions of all theoretical results?
    \answerNA{We do not include theoretical results.}
        \item Did you include complete proofs of all theoretical results?
    \answerNA{We do not include theoretical results.}
\end{enumerate}

\item If you ran experiments...
\begin{enumerate}
  \item Did you include the code, data, and instructions needed to reproduce the main experimental results (either in the supplemental material or as a URL)?
    \answerYes{See the supplementary material.}
  \item Did you specify all the training details (e.g., data splits, hyperparameters, how they were chosen)?
    \answerYes{The supplementary material includes the script used to create data splits. The hyperparameters are either explicitly described in \autoref{sec:implementation-details} and supplementary material, or tuned as described in \autoref{sec:implementation-details}}.
        \item Did you report error bars (e.g., with respect to the random seed after running experiments multiple times)?
    \answerYes{We provide standard deviations in the supplementary material, see \autoref{tab:A_single_models} and see \autoref{tab:A_ensembles}}
        \item Did you include the total amount of compute and the type of resources used (e.g., type of GPUs, internal cluster, or cloud provider)?
    \answerYes{The experiment reports included in the supplementary material provide the information about the used hardware and execution times.}
\end{enumerate}

\item If you are using existing assets (e.g., code, data, models) or curating/releasing new assets...
\begin{enumerate}
  \item If your work uses existing assets, did you cite the creators?
    \answerYes{See \autoref{A:datasets}.}
  \item Did you mention the license of the assets?
    \answerYes{In the \texttt{README.md} file in the supplementary material, we refer to the original licenses of the used datasets.}
  \item Did you include any new assets either in the supplemental material or as a URL?
    \answerNA{We do not provide new datasets.}
  \item Did you discuss whether and how consent was obtained from people whose data you're using/curating?
    \answerNA{We use publicly available datasets.}
  \item Did you discuss whether the data you are using/curating contains personally identifiable information or offensive content?
    \answerNA{We use publicly available datasets.}
\end{enumerate}

\item If you used crowdsourcing or conducted research with human subjects...
\begin{enumerate}
  \item Did you include the full text of instructions given to participants and screenshots, if applicable?
    \answerNA{We did not use crowdsourcing. We did not conduct research with human subjects.}
  \item Did you describe any potential participant risks, with links to Institutional Review Board (IRB) approvals, if applicable?
    \answerNA{We did not use crowdsourcing. We did not conduct research with human subjects.}
  \item Did you include the estimated hourly wage paid to participants and the total amount spent on participant compensation?
    \answerNA{We did not use crowdsourcing. We did not conduct research with human subjects.}
\end{enumerate}

\end{enumerate}


\newpage
\appendix

\section*{Supplementary material}

\section{MLP with embeddings for numerical features}
\label{A:mlp-with-embeddings}

We provide visual explanation of how embeddings are passed to MLP in \autoref{fig:A_mlp-without-embeddings} and \autoref{fig:A_mlp-with-embeddings}.
Also, we provide the formal explanation in \autoref{eq:A_mlp-with-embeddings} (categorical features are omitted for simplicity).

\begin{minipage}{0.39\textwidth}
    \centering
    \includegraphics[width=0.85\linewidth]{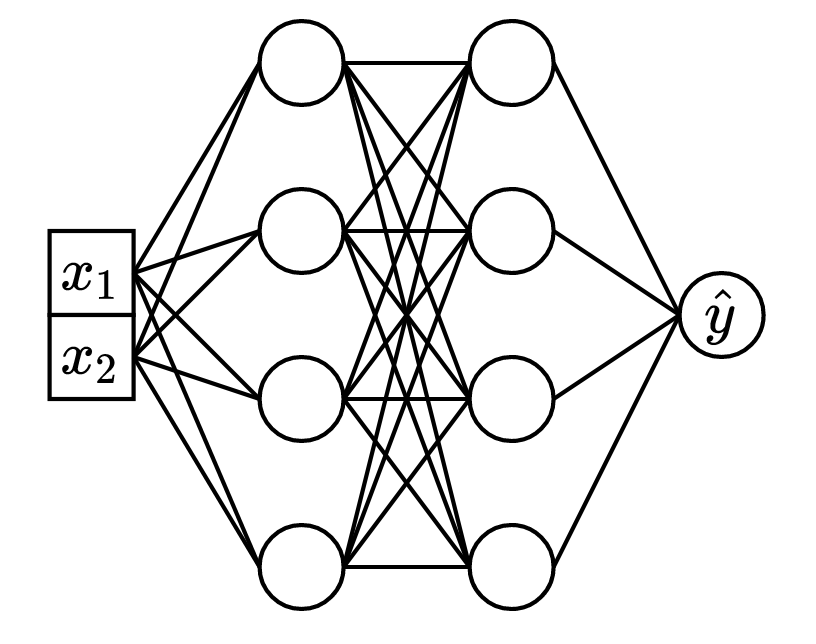}
    \captionof{figure}{The vanilla MLP. The model takes two numerical features as input.}
    \label{fig:A_mlp-without-embeddings}
\end{minipage}
\hspace{0.05\linewidth}
\begin{minipage}{0.51\textwidth}
    \vspace{-1em}
    \centering
    \includegraphics[width=0.85\linewidth]{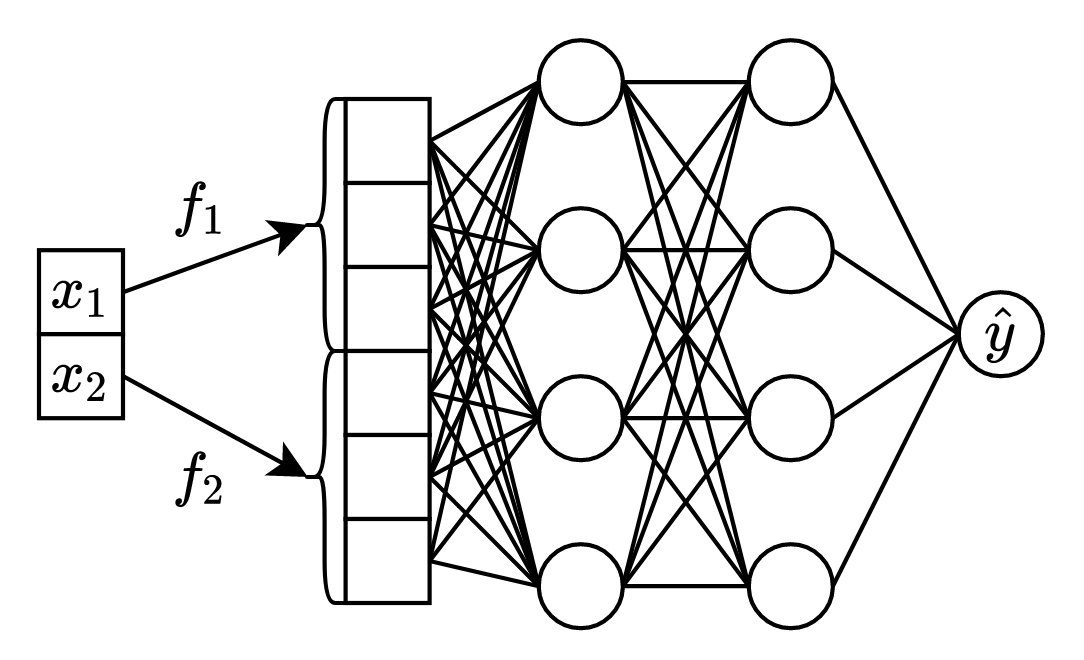}
    \captionof{figure}{The same MLP as in \autoref{fig:A_mlp-without-embeddings}, but now with embeddings for numerical features.}
    \label{fig:A_mlp-with-embeddings}
\end{minipage}

\begin{equation}
\label{eq:A_mlp-with-embeddings}
    \MLP(z_1,\ \dots,\ z_k) = \MLP\left(\texttt{concat}[z_1,\ \dots,\ z_k]\right)
    \qquad
    \texttt{concat}[z_1,\ \dots,\ z_k] \in \R^{d_1 + \dots\ + d_k}
\end{equation}

\section{Target-aware piecewise linear encoding}

\begin{figure}[h]
    \centering
    \includegraphics[width=0.7\linewidth]{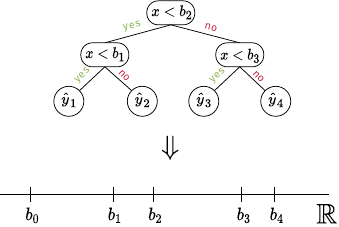}
    \caption{Obtaining bins for \PLE\ from decision trees.}
    \label{fig:A_target-aware-ple}
\end{figure}

We provide visualisation of target-aware \PLE\ (\autoref{sec:target-aware-ple}) in \autoref{fig:A_target-aware-ple}.

\newpage
\section{Additional details on datasets}
\label{A:datasets}

\begin{table}[h]
    \setlength\tabcolsep{2.2pt}
    \centering
    \caption{Details on datasets, used for experiments}
    \label{tab:A_datasets}
    \vspace{1em}
    \scalebox{0.95}{{\footnotesize \begin{tabular}{lcccccccc}
\toprule
Abbr & Name & \# Train & \# Validation & \# Test & \# Num & \# Cat & Task type & Batch size \\
\midrule
GE & Gesture Phase & $6318$ & $1580$ & $1975$ & $32$ & $0$ & Multiclass & 128 \\
CH & Churn Modelling & $6400$ & $1600$ & $2000$ & $10$ & $1$ & Binclass & 128 \\
CA & California Housing & $13209$ & $3303$ & $4128$ & $8$ & $0$ & Regression & 256 \\
HO & House 16H & $14581$ & $3646$ & $4557$ & $16$ & $0$ & Regression & 256 \\
AD & Adult & $26048$ & $6513$ & $16281$ & $6$ & $8$ & Binclass & 256 \\
OT & Otto Group Products & $39601$ & $9901$ & $12376$ & $93$ & $0$ & Multiclass & 512 \\
HI & Higgs Small & $62751$ & $15688$ & $19610$ & $28$ & $0$ & Binclass & 512 \\
FB & Facebook Comments Volume & $157638$ & $19722$ & $19720$ & $50$ & $1$ & Regression & 512 \\
SA & Santander Customer Transactions & $128000$ & $32000$ & $40000$ & $200$ & $0$ & Binclass & 1024 \\
CO & Covertype & $371847$ & $92962$ & $116203$ & $54$ & $0$ & Multiclass & 1024 \\
MI & MSLR-WEB10K (Fold 1) & $723412$ & $235259$ & $241521$ & $136$ & $0$ & Regression & 1024 \\
\bottomrule
\end{tabular}}}
\end{table}

We used the following datasets:
\begin{itemize}[nosep, leftmargin=2em]
    \item Gesture Phase Prediction (\citet{gesture})
    \item Churn Modeling\footnote{https://www.kaggle.com/shrutimechlearn/churn-modelling}
    \item California Housing (real estate data, \citet{california})
    \item House 16H\footnote{https://www.openml.org/d/574}
    \item Adult (income estimation, \citet{adult})
    \item Otto Group Product Classification\footnote{https://www.kaggle.com/c/otto-group-product-classification-challenge/data}
    \item Higgs (simulated physical particles, \citet{higgs}; we use the version with 98K samples available in the OpenML repository \citep{openml})
    \item Santander Customer Transaction Prediction\footnote{https://www.kaggle.com/c/santander-customer-transaction-prediction}
    \item Facebook Comments (\citet{fb-comments})
    \item Covertype (forest characteristics, \citet{covertype})
    \item Microsoft (search queries, \citet{microsoft}). We follow the pointwise approach to learning-to-rank and treat this ranking problem as a regression problem.
\end{itemize}

\section{Additional analysis}

\subsection{Testing quantile-based \PLE\ on the synthetic GBDT-friendly dataset}
\label{A:synthetic}

\begin{figure}[h]
    \centering
    \includegraphics[width=0.7\linewidth]{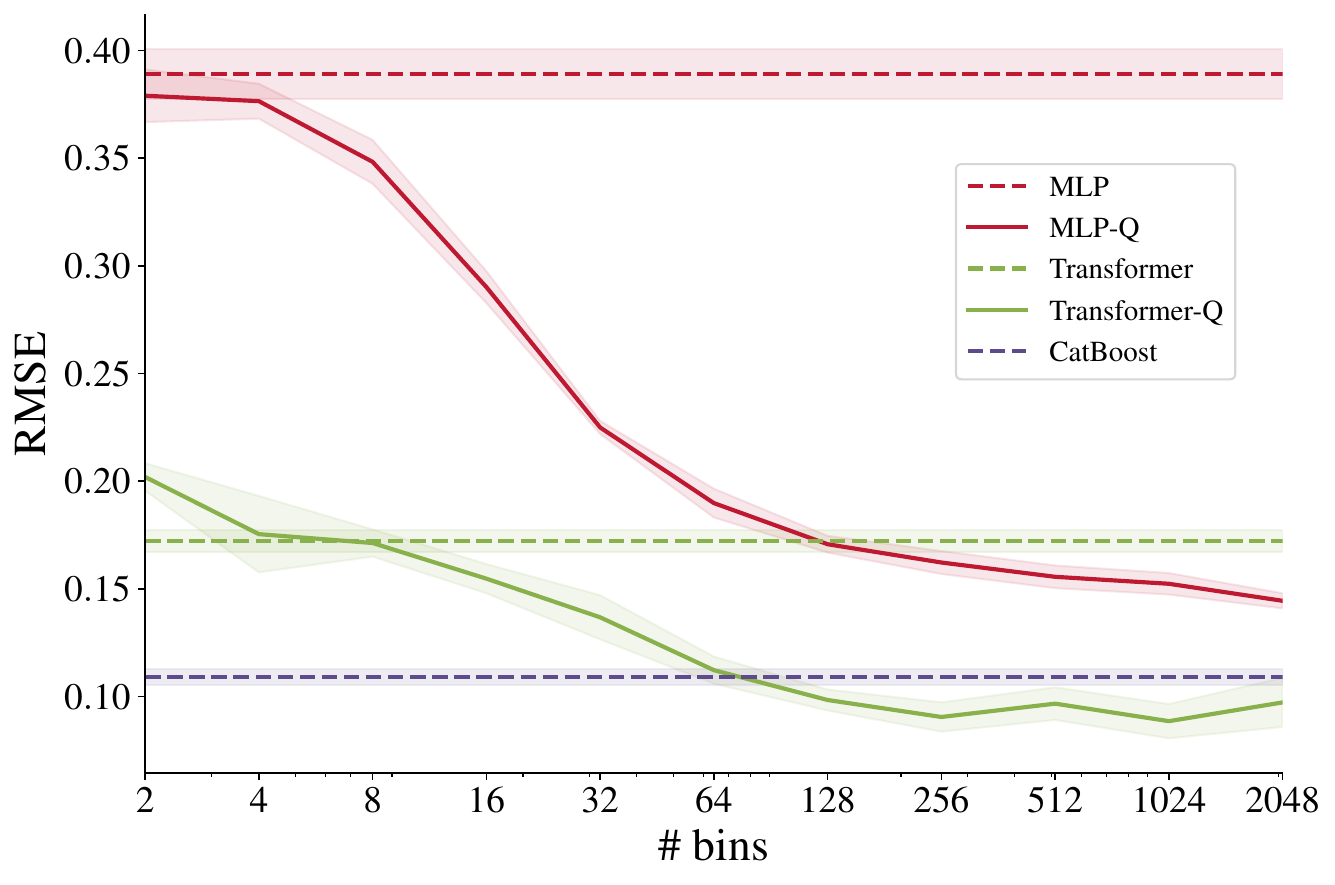}
    \caption{RMSE (averaged over five random seeds) of different approaches on the same synthetic GBDT-friendly task.
    Using \PLE-representations (``-Q'') instead of scalar values improves the performance of MLP and Transformer.
    Note that in practice, increasing the number of bins does not always lead to better results.
    }
    \label{fig:A_synthetic}
\end{figure}

In this section, we apply the quantile-based piecewise linear encoding (described in \autoref{sec:ple-quantiles} to MLP and Transformer on the synthetic GBDT-friendly dataset described in section 5.1 in \citet{revisiting}.
In a nutshell, features of this dataset are sampled randomly from $\mathcal{N}(0, 1)$, and the target is produced by an ensemble of randomly constructed decision trees applied to the sampled features.
This task turns out to be easy for GBDT, but hard for traditional DL models \citep{revisiting}.
The results are visualized in \autoref{fig:A_synthetic}.
As the plot shows, \PLE-representations can be helpful for both MLP and Transformer backbones. In the considered synthetic setup, increasing the number of bins leads to better results, however, in practice, using too many bins can lead to overfitting; therefore, we recommend tuning the number of bins based on a validation set.

\textbf{Technical details.} Our dataset has $10,000$ objects, $8$ features and the target was produced by $16$ decision trees of depth $6$.
CatBoost is trained with the default hyperparameters.
Transformer is trained with the default hyperparameters of FT-Transformer.
The MLP backbone has four layers of size 256 each.
Importantly, the task GBDT-friendly, which can be illustrated by the performance of the \textit{tuned} MLP: $0.2229 \pm 0.0055$ (it is still worse than the performance of CatBoost).
The remaining details can be found in the source code.

\subsection{Fourier features}
\label{A:fourier}

In this section, we test Fourier features implemented exactly as in \citet{fourier-features}, i.e. pre-activation coefficients are not trained and features are mixed right from the start.
Importantly, the latter means that this approach is not covered by the embedding framework described in \autoref{sec:general-framework}.
As reported in \autoref{tab:A_fourier}, MLP equipped with the original Fourier features does not perform well even compared to the vanilla MLP.
So, it seems to be important to embed each feature separately as described in \autoref{sec:general-framework}.

\begin{table}[h]
    \setlength\tabcolsep{2.2pt}
    \centering
    \caption{Results for the vanilla MLP and MLP equipped with Fourier features \citep{fourier-features}. Notation
follows Table 3 and Table 2.}
    \label{tab:A_fourier}
    \vspace{1em}
    \scalebox{0.9}{{\footnotesize \begin{tabular}{lccccccccccc}
\toprule
{} & GE \textuparrow & CH \textuparrow & CA \textdownarrow & HO \textdownarrow & AD \textuparrow & OT \textuparrow & HI \textuparrow & FB \textdownarrow & SA \textuparrow & CO \textuparrow & MI \textdownarrow \\
\midrule
MLP             & $\mathbf{0.632}$ & $\mathbf{0.856}$ & $\mathbf{0.495}$ & $\mathbf{3.204}$ & $0.854$ & $\mathbf{0.818}$ & $\mathbf{0.720}$ & $\mathbf{5.686}$ & $0.912$ & $\mathbf{0.964}$ & $\mathbf{0.747}$ \\
MLP (Fourier features) & $0.612$ & $0.845$ & $\mathbf{0.495}$ & $3.267$ & $\mathbf{0.858}$ & $0.810$ & $0.711$ & $5.767$ & $\mathbf{0.915}$ & $0.961$ & $0.749$ \\
\bottomrule
\end{tabular}}}
\end{table}


\section{Implementation details}
\label{A:implementation-details}

We mostly follow \citet{revisiting} in terms of the tuning, training and evaluation protocols.

\textbf{Data preprocessing}.
Preliminary data preprocessing is known to be crucial for the optimization of tabular DL models.
For each dataset, the same preprocessing was used for all deep models for a fair comparison.
For all datasets except for Otto Group Product Classification, we use the quantile transformation from the Scikit-learn library \citep{scikit-learn}.
For Otto Group Product Classification, we do not apply any feature preprocessing.
We also apply standardization to regression targets for all algorithms.

\textbf{Tuning}.
For every dataset, we carefully tune each model's hyperparameters.
The best hyperparameters are the ones that perform best on the validation set, so the test set is never used for tuning.
For most algorithms, we use the Optuna library \citep{optuna} to run Bayesian optimization (the Tree-Structured Parzen Estimator algorithm), which is reported to be superior to random search \citep{hp-tuning}. The search spaces for all hyperparameters are reported in the appendix.

\textbf{Evaluation}. For each tuned configuration, we run 15 experiments with different random seeds and report the average performance on the test set.

\textbf{Ensembles}. For each model-dataset pair, we obtain three ensembles by splitting the 15 single models into three disjoint groups of equal size and averaging predictions of single models within each group.

\textbf{Neural networks}.
The implementations of the MLP, ResNet, and Transformer backbones are taken from \citet{revisiting}.
We minimize cross-entropy for classification problems and mean squared error for regression problems.
We use the AdamW optimizer \citep{adamw}.
We do not apply learning rate schedules.
For each dataset, we use a predefined batch size (see \autoref{A:datasets} for the specific values).
We continue training until there are $\texttt{patience} + 1$ consecutive epochs without improvements on the validation set; we set $\texttt{patience} = 16$ for all models.

\textbf{Categorical features.}
For CatBoost, we employ the built-in support for categorical features. For all other algorithms, we use the one-hot encoding.

\subsection{One-blob encoding}
\label{A:one-blob}

In \autoref{sec:ablation}, we used a slightly generalized version of the original one-blob encoding \citep{one-blob}.
Namely, while the original sets the width of the kernel to $T^{-1}$ ($T$ is the number of bins), we set it to $T^{-\gamma}$ and tune $\gamma$.

\subsection{Hyperparameter tuning configurations}
\label{A:hyperparameters}

\subsection{CatBoost}
\vspace{1em}
We fix and do not tune the following hyperparameters:
\begin{itemize}
    \itemsep0em
    \item $\texttt{early-stopping-rounds} = 50$
    \item $\texttt{od-pval} = 0.001$
    \item $\texttt{iterations} = 2000$
\end{itemize}

For tuning on the MI and CO datasets, we set the \texttt{task\_type} parameter to ``GPU''. In all other cases (including the evaluation on these two datasets), we set this parameter to ``CPU''.

\begin{table}[H]
\centering
\caption{CatBoost hyperparameter space}
\label{tab:A-catboost-space}
\vspace{1em}
{\renewcommand{\arraystretch}{1.2}
\begin{tabular}{ll}
    \toprule
    Parameter & Distribution \\
    \midrule
    Max depth & $\mathrm{UniformInt[1,10]}$ \\
    Learning rate & $\mathrm{LogUniform}[0.001, 1]$ \\
    Bagging temperature &  $\mathrm{Uniform}[0, 1]$ \\
    L2 leaf reg  & $\mathrm{LogUniform}[1, 10]$ \\
    Leaf estimation iterations &  $\mathrm{UniformInt}[1, 10]$ \\
    \midrule
    \# Iterations & 100 \\
    \bottomrule
\end{tabular}}
\end{table}

\subsection{XGBoost}

We fix and do not tune the following hyperparameters:
\begin{itemize}
    \itemsep0em
    \item $\texttt{booster} = \text{"gbtree"}$
    \item $\texttt{early-stopping-rounds} = 50$
    \item $\texttt{n-estimators} = 2000$
\end{itemize}

\begin{table}[H]
\centering
\caption{XGBoost hyperparameter space.}
\label{tab:A-xgboost-space}
\vspace{1em}
{\renewcommand{\arraystretch}{1.2}
\begin{tabular}{ll}
    \toprule
    Parameter & Distribution \\
    \midrule
    Max depth & $\mathrm{UniformInt[3,10]}$ \\
    Min child weight & $\mathrm{LogUniform}[0.0001, 100]$ \\
    Subsample & $\mathrm{Uniform}[0.5, 1]$ \\
    Learning rate & $\mathrm{LogUniform}[0.001, 1]$ \\
    Col sample by tree & $\mathrm{Uniform}[0.5, 1]$ \\ 
    Gamma & $\{0, \mathrm{LogUniform}[0.001, 100]\}$ \\ 
    Lambda & $\{0, \mathrm{LogUniform}[0.1, 10]\}$ \\ 
    \midrule
    \# Iterations & 100 \\
    \bottomrule
\end{tabular}}
\end{table}

\subsection{MLP}

\begin{table}[H]
\centering
\caption{MLP hyperparameter space. }
\label{tab:A-mlp-space}
\vspace{1em}
{\renewcommand{\arraystretch}{1.2}
\begin{tabular}{ll}
    \toprule
    Parameter & Distribution \\
    \midrule
    \# Layers & $\mathrm{UniformInt}[1,16]$ \\
    Layer size & $\mathrm{UniformInt}[1,1024]$ \\
    Dropout &   $\{0, \mathrm{Uniform}[0, 0.5]\}$ \\
    Learning rate &  $\mathrm{LogUniform}[5e\text{-}5, 0.005]$ \\
    Weight decay &  $\{0, \mathrm{LogUniform}[1e\text{-}6, 1e\text{-}3] \}$ \\
    \midrule
    \# Iterations & 100 \\
    \bottomrule
\end{tabular}}
\end{table}

\subsection{ResNet}

\begin{table}[H]
\centering
\caption{ResNet hyperparameter space.}
\label{tab:A-resnet-space}
\vspace{1em}
{\renewcommand{\arraystretch}{1.2}
\begin{tabular}{ll}
    \toprule
    Parameter & Distribution \\
    \midrule
    \# Layers & $\mathrm{UniformInt}[1,8]$ \\
    Layer size & $\mathrm{UniformInt}[32,512]$ \\
    Hidden factor & $\mathrm{Uniform}[1, 4]$\\
    Hidden dropout &  $\mathrm{Uniform}[0, 0.5]$ \\
    Residual dropout & $\{0, \mathrm{Uniform}[0, 0.5]\}$ \\
    Learning rate & $\mathrm{LogUniform}[5e\text{-}5, 0.005]$ \\
    Weight decay & $\{0, \mathrm{LogUniform}[1e\text{-}6, 1e\text{-}3] \}$ \\
    \midrule
    \# Iterations & 100 \\
    \bottomrule
\end{tabular}}
\end{table}

\subsection{Transformer}

\begin{table}[H]
\centering
\caption{Transformer hyperparameter space. Here (A) = \{SA, CO, MI\} and (B) = the rest }
\label{tab:S-transformer-space}
\vspace{1em}
{\renewcommand{\arraystretch}{1.2}
\begin{tabular}{ll}
    \toprule
    Parameter & (Datasets) Distribution \\
    \midrule
    \# Layers & (A) $\mathrm{UniformInt}[2,4]$, (B) $\mathrm{UniformInt}[1,4]$ \\
    Embedding size & (A) $\mathrm{UniformInt}[192,512]$, (B) $\mathrm{UniformInt}[96,512]$ \\
    Residual dropout &  (A) $\mathrm{Const}(0.0)$, (B) $\{0, \mathrm{Uniform}[0, 0.2] \}$ \\
    Attention dropout &  (A,B) $\mathrm{Uniform}[0, 0.5]$ \\
    FFN dropout & (A,B) $\mathrm{Uniform}[0, 0.5]$ \\
    FFN factor & (A,B) $\mathrm{Uniform}[\nicefrac{2}{3}, \nicefrac{8}{3}]$ \\
    Learning rate & (A) $\mathrm{LogUniform}[1e\text{-}5, 3e\text{-}4]$, (B) $\mathrm{LogUniform}[1e\text{-}5, 1e\text{-}3]$ \\
    Weight decay & (A) $\mathrm{Const}(1e\text{-}5)$, (B) $\mathrm{LogUniform}[1e\text{-}6, 1e\text{-}4]$ \\
    \midrule
    \# Iterations & (A) 50, (B) 100 \\
    \bottomrule
\end{tabular}}
\end{table}

\subsection{Embedding hyperparameters}

The distribution for the output dimensions of linear layers is $\mathrm{UniformInt}[1, 128]$.

\texttt{PLE}. We share the same hyperparameter space for \PLE\ across all datasets and models.
For the quantile-based \PLE, the distribution for the number of quantiles is $\mathrm{UniformInt}[2,256]$.
For the target-aware (tree-based) \PLE, the distribution for the number of leaves is $\mathrm{UniformInt}[2,256]$, the distribution for the minimum number of items per leaf is $\mathrm{UniformInt}[1,128]$ and the distribution for the minimum information gain required for making a split is $\mathrm{LogUniform}[1e\text{-}9, 0.01]$.

\texttt{Periodic}. The distribution for $k$ (see \autoref{eq:periodic}) is $\mathrm{UniformInt}[1,128]$.

\section{Extended tables with experimental results}
\label{A:stds}

The scores with standard deviations for single models and ensembles are provided in \autoref{tab:A_single_models} and \autoref{tab:A_ensembles} respectively.
Please, refer to \autoref{tab:names} to learn about the model names.

Additionally, we include the results for the DICE embeddings \citep{dice}, which is a general way to represent numbers with vectors introduced in the context of NLP.
The results though demonstrate that it is a suboptimal approach in tabular data problems.
 
\begin{sidewaystable}[t!]
    \setlength\tabcolsep{2.2pt}
    \centering
    \caption{Extended results for single models}
    \label{tab:A_single_models}
    \vspace{1em}
    {\scriptsize \begin{tabular}{lccccccccccc}
\toprule
{} & GE \textuparrow & CH \textuparrow & CA \textdownarrow & HO \textdownarrow & AD \textuparrow & OT \textuparrow & HI \textuparrow & FB \textdownarrow & SA \textuparrow & CO \textuparrow & MI \textdownarrow \\
\midrule
CatBoost        & $0.683 \scriptscriptstyle \pm \scriptstyle 4.7e\text{-}3$ & $0.861 \scriptscriptstyle \pm \scriptstyle 3.5e\text{-}3$ & $0.433 \scriptscriptstyle \pm \scriptstyle 1.8e\text{-}3$ & $3.115 \scriptscriptstyle \pm \scriptstyle 1.9e\text{-}2$ & $0.872 \scriptscriptstyle \pm \scriptstyle 9.0e\text{-}4$ & $0.824 \scriptscriptstyle \pm \scriptstyle 1.1e\text{-}3$ & $0.726 \scriptscriptstyle \pm \scriptstyle 1.0e\text{-}3$ & $5.324 \scriptscriptstyle \pm \scriptstyle 4.1e\text{-}2$ & $0.923 \scriptscriptstyle \pm \scriptstyle 3.6e\text{-}4$ & $0.966 \scriptscriptstyle \pm \scriptstyle 3.3e\text{-}4$ & $0.743 \scriptscriptstyle \pm \scriptstyle 3.1e\text{-}4$ \\
XGBoost         & $0.678 \scriptscriptstyle \pm \scriptstyle 4.9e\text{-}3$ & $0.858 \scriptscriptstyle \pm \scriptstyle 2.2e\text{-}3$ & $0.436 \scriptscriptstyle \pm \scriptstyle 2.5e\text{-}3$ & $3.160 \scriptscriptstyle \pm \scriptstyle 6.9e\text{-}3$ & $0.874 \scriptscriptstyle \pm \scriptstyle 8.2e\text{-}4$ & $0.825 \scriptscriptstyle \pm \scriptstyle 2.3e\text{-}3$ & $0.724 \scriptscriptstyle \pm \scriptstyle 1.0e\text{-}3$ & $5.383 \scriptscriptstyle \pm \scriptstyle 2.9e\text{-}2$ & $0.918 \scriptscriptstyle \pm \scriptstyle 5.0e\text{-}3$ & $0.969 \scriptscriptstyle \pm \scriptstyle 6.1e\text{-}4$ & $0.742 \scriptscriptstyle \pm \scriptstyle 1.6e\text{-}4$ \\
\midrule
MLP             & $0.632 \scriptscriptstyle \pm \scriptstyle 1.4e\text{-}2$ & $0.856 \scriptscriptstyle \pm \scriptstyle 2.8e\text{-}3$ & $0.495 \scriptscriptstyle \pm \scriptstyle 4.3e\text{-}3$ & $3.204 \scriptscriptstyle \pm \scriptstyle 4.0e\text{-}2$ & $0.854 \scriptscriptstyle \pm \scriptstyle 1.6e\text{-}3$ & $0.818 \scriptscriptstyle \pm \scriptstyle 3.1e\text{-}3$ & $0.720 \scriptscriptstyle \pm \scriptstyle 2.3e\text{-}3$ & $5.686 \scriptscriptstyle \pm \scriptstyle 4.7e\text{-}2$ & $0.912 \scriptscriptstyle \pm \scriptstyle 4.3e\text{-}4$ & $0.964 \scriptscriptstyle \pm \scriptstyle 8.6e\text{-}4$ & $0.747 \scriptscriptstyle \pm \scriptstyle 2.5e\text{-}4$ \\
MLP-L           & $0.639 \scriptscriptstyle \pm \scriptstyle 1.3e\text{-}2$ & $0.861 \scriptscriptstyle \pm \scriptstyle 2.1e\text{-}3$ & $0.475 \scriptscriptstyle \pm \scriptstyle 5.4e\text{-}3$ & $3.123 \scriptscriptstyle \pm \scriptstyle 4.5e\text{-}2$ & $0.856 \scriptscriptstyle \pm \scriptstyle 1.6e\text{-}3$ & $0.820 \scriptscriptstyle \pm \scriptstyle 1.5e\text{-}3$ & $0.723 \scriptscriptstyle \pm \scriptstyle 1.6e\text{-}3$ & $5.684 \scriptscriptstyle \pm \scriptstyle 4.5e\text{-}2$ & $0.916 \scriptscriptstyle \pm \scriptstyle 3.5e\text{-}4$ & $0.963 \scriptscriptstyle \pm \scriptstyle 9.3e\text{-}4$ & $0.748 \scriptscriptstyle \pm \scriptstyle 4.1e\text{-}4$ \\
MLP-LR          & $0.642 \scriptscriptstyle \pm \scriptstyle 1.5e\text{-}2$ & $0.860 \scriptscriptstyle \pm \scriptstyle 3.0e\text{-}3$ & $0.471 \scriptscriptstyle \pm \scriptstyle 2.6e\text{-}3$ & $3.084 \scriptscriptstyle \pm \scriptstyle 2.7e\text{-}2$ & $0.857 \scriptscriptstyle \pm \scriptstyle 1.9e\text{-}3$ & $0.819 \scriptscriptstyle \pm \scriptstyle 1.6e\text{-}3$ & $0.726 \scriptscriptstyle \pm \scriptstyle 1.9e\text{-}3$ & $5.625 \scriptscriptstyle \pm \scriptstyle 5.6e\text{-}2$ & $0.923 \scriptscriptstyle \pm \scriptstyle 3.1e\text{-}4$ & $0.963 \scriptscriptstyle \pm \scriptstyle 1.4e\text{-}3$ & $0.746 \scriptscriptstyle \pm \scriptstyle 3.9e\text{-}4$ \\
MLP-LRLR        & $0.654 \scriptscriptstyle \pm \scriptstyle 1.7e\text{-}2$ & $0.861 \scriptscriptstyle \pm \scriptstyle 2.6e\text{-}3$ & $0.460 \scriptscriptstyle \pm \scriptstyle 3.8e\text{-}3$ & $3.070 \scriptscriptstyle \pm \scriptstyle 3.6e\text{-}2$ & $0.857 \scriptscriptstyle \pm \scriptstyle 1.4e\text{-}3$ & $0.819 \scriptscriptstyle \pm \scriptstyle 2.2e\text{-}3$ & $0.725 \scriptscriptstyle \pm \scriptstyle 1.2e\text{-}3$ & $5.551 \scriptscriptstyle \pm \scriptstyle 4.6e\text{-}2$ & $0.923 \scriptscriptstyle \pm \scriptstyle 3.0e\text{-}4$ & $0.963 \scriptscriptstyle \pm \scriptstyle 1.4e\text{-}3$ & $0.746 \scriptscriptstyle \pm \scriptstyle 3.0e\text{-}4$ \\
MLP-Q           & $0.653 \scriptscriptstyle \pm \scriptstyle 8.9e\text{-}3$ & $0.854 \scriptscriptstyle \pm \scriptstyle 3.0e\text{-}3$ & $0.464 \scriptscriptstyle \pm \scriptstyle 3.1e\text{-}3$ & $3.163 \scriptscriptstyle \pm \scriptstyle 3.1e\text{-}2$ & $0.859 \scriptscriptstyle \pm \scriptstyle 1.6e\text{-}3$ & $0.816 \scriptscriptstyle \pm \scriptstyle 2.6e\text{-}3$ & $0.721 \scriptscriptstyle \pm \scriptstyle 1.0e\text{-}3$ & $5.766 \scriptscriptstyle \pm \scriptstyle 5.3e\text{-}2$ & $0.922 \scriptscriptstyle \pm \scriptstyle 6.3e\text{-}4$ & $0.968 \scriptscriptstyle \pm \scriptstyle 6.9e\text{-}4$ & $0.750 \scriptscriptstyle \pm \scriptstyle 3.7e\text{-}4$ \\
MLP-Q-LR        & $0.646 \scriptscriptstyle \pm \scriptstyle 6.3e\text{-}3$ & $0.857 \scriptscriptstyle \pm \scriptstyle 2.6e\text{-}3$ & $0.455 \scriptscriptstyle \pm \scriptstyle 3.4e\text{-}3$ & $3.184 \scriptscriptstyle \pm \scriptstyle 3.1e\text{-}2$ & $0.863 \scriptscriptstyle \pm \scriptstyle 1.7e\text{-}3$ & $0.811 \scriptscriptstyle \pm \scriptstyle 1.8e\text{-}3$ & $0.720 \scriptscriptstyle \pm \scriptstyle 1.5e\text{-}3$ & $5.394 \scriptscriptstyle \pm \scriptstyle 1.5e\text{-}1$ & $0.923 \scriptscriptstyle \pm \scriptstyle 6.1e\text{-}4$ & $0.969 \scriptscriptstyle \pm \scriptstyle 4.8e\text{-}4$ & $0.747 \scriptscriptstyle \pm \scriptstyle 3.9e\text{-}4$ \\
MLP-Q-LRLR      & $0.644 \scriptscriptstyle \pm \scriptstyle 6.2e\text{-}3$ & $0.859 \scriptscriptstyle \pm \scriptstyle 2.2e\text{-}3$ & $0.452 \scriptscriptstyle \pm \scriptstyle 4.3e\text{-}3$ & $3.118 \scriptscriptstyle \pm \scriptstyle 4.6e\text{-}2$ & $0.869 \scriptscriptstyle \pm \scriptstyle 1.5e\text{-}3$ & $0.812 \scriptscriptstyle \pm \scriptstyle 2.5e\text{-}3$ & $0.724 \scriptscriptstyle \pm \scriptstyle 1.3e\text{-}3$ & $5.618 \scriptscriptstyle \pm \scriptstyle 2.0e\text{-}1$ & $0.924 \scriptscriptstyle \pm \scriptstyle 4.5e\text{-}4$ & $0.969 \scriptscriptstyle \pm \scriptstyle 1.2e\text{-}3$ & $0.748 \scriptscriptstyle \pm \scriptstyle 4.6e\text{-}4$ \\
MLP-T           & $0.647 \scriptscriptstyle \pm \scriptstyle 5.7e\text{-}3$ & $0.861 \scriptscriptstyle \pm \scriptstyle 1.1e\text{-}3$ & $0.447 \scriptscriptstyle \pm \scriptstyle 2.0e\text{-}3$ & $3.149 \scriptscriptstyle \pm \scriptstyle 5.2e\text{-}2$ & $0.864 \scriptscriptstyle \pm \scriptstyle 6.3e\text{-}4$ & $0.821 \scriptscriptstyle \pm \scriptstyle 1.8e\text{-}3$ & $0.720 \scriptscriptstyle \pm \scriptstyle 1.9e\text{-}3$ & $5.577 \scriptscriptstyle \pm \scriptstyle 3.7e\text{-}2$ & $0.923 \scriptscriptstyle \pm \scriptstyle 3.0e\text{-}4$ & $0.967 \scriptscriptstyle \pm \scriptstyle 1.1e\text{-}3$ & $0.749 \scriptscriptstyle \pm \scriptstyle 4.4e\text{-}4$ \\
MLP-T-LR        & $0.640 \scriptscriptstyle \pm \scriptstyle 6.9e\text{-}3$ & $0.861 \scriptscriptstyle \pm \scriptstyle 2.0e\text{-}3$ & $0.439 \scriptscriptstyle \pm \scriptstyle 3.7e\text{-}3$ & $3.207 \scriptscriptstyle \pm \scriptstyle 5.2e\text{-}2$ & $0.868 \scriptscriptstyle \pm \scriptstyle 1.1e\text{-}3$ & $0.818 \scriptscriptstyle \pm \scriptstyle 1.3e\text{-}3$ & $0.724 \scriptscriptstyle \pm \scriptstyle 1.7e\text{-}3$ & $5.508 \scriptscriptstyle \pm \scriptstyle 3.0e\text{-}2$ & $0.924 \scriptscriptstyle \pm \scriptstyle 2.4e\text{-}4$ & $0.968 \scriptscriptstyle \pm \scriptstyle 7.2e\text{-}4$ & $0.747 \scriptscriptstyle \pm \scriptstyle 5.7e\text{-}4$ \\
MLP-T-LRLR      & $0.629 \scriptscriptstyle \pm \scriptstyle 1.0e\text{-}2$ & $0.857 \scriptscriptstyle \pm \scriptstyle 2.4e\text{-}3$ & $0.446 \scriptscriptstyle \pm \scriptstyle 3.6e\text{-}3$ & $3.153 \scriptscriptstyle \pm \scriptstyle 4.0e\text{-}2$ & $0.870 \scriptscriptstyle \pm \scriptstyle 9.9e\text{-}4$ & $0.818 \scriptscriptstyle \pm \scriptstyle 2.1e\text{-}3$ & $0.725 \scriptscriptstyle \pm \scriptstyle 1.3e\text{-}3$ & $5.553 \scriptscriptstyle \pm \scriptstyle 2.4e\text{-}2$ & $0.924 \scriptscriptstyle \pm \scriptstyle 3.6e\text{-}4$ & $0.967 \scriptscriptstyle \pm \scriptstyle 8.5e\text{-}4$ & $0.748 \scriptscriptstyle \pm \scriptstyle 5.8e\text{-}4$ \\
MLP-P           & $0.631 \scriptscriptstyle \pm \scriptstyle 1.7e\text{-}2$ & $0.860 \scriptscriptstyle \pm \scriptstyle 3.1e\text{-}3$ & $0.489 \scriptscriptstyle \pm \scriptstyle 2.4e\text{-}3$ & $3.129 \scriptscriptstyle \pm \scriptstyle 4.3e\text{-}2$ & $0.869 \scriptscriptstyle \pm \scriptstyle 1.5e\text{-}3$ & $0.807 \scriptscriptstyle \pm \scriptstyle 4.3e\text{-}3$ & $0.723 \scriptscriptstyle \pm \scriptstyle 1.5e\text{-}3$ & $5.845 \scriptscriptstyle \pm \scriptstyle 6.4e\text{-}2$ & $0.923 \scriptscriptstyle \pm \scriptstyle 4.3e\text{-}4$ & $0.968 \scriptscriptstyle \pm \scriptstyle 9.0e\text{-}4$ & $0.747 \scriptscriptstyle \pm \scriptstyle 3.1e\text{-}4$ \\
MLP-PL          & $0.641 \scriptscriptstyle \pm \scriptstyle 1.0e\text{-}2$ & $0.859 \scriptscriptstyle \pm \scriptstyle 2.4e\text{-}3$ & $0.467 \scriptscriptstyle \pm \scriptstyle 2.9e\text{-}3$ & $3.113 \scriptscriptstyle \pm \scriptstyle 3.1e\text{-}2$ & $0.868 \scriptscriptstyle \pm \scriptstyle 1.1e\text{-}3$ & $0.819 \scriptscriptstyle \pm \scriptstyle 1.7e\text{-}3$ & $0.727 \scriptscriptstyle \pm \scriptstyle 1.7e\text{-}3$ & $5.530 \scriptscriptstyle \pm \scriptstyle 9.5e\text{-}2$ & $0.924 \scriptscriptstyle \pm \scriptstyle 4.0e\text{-}4$ & $0.969 \scriptscriptstyle \pm \scriptstyle 5.0e\text{-}4$ & $0.746 \scriptscriptstyle \pm \scriptstyle 2.6e\text{-}4$ \\
MLP-PLR         & $0.674 \scriptscriptstyle \pm \scriptstyle 1.0e\text{-}2$ & $0.857 \scriptscriptstyle \pm \scriptstyle 2.4e\text{-}3$ & $0.467 \scriptscriptstyle \pm \scriptstyle 5.8e\text{-}3$ & $3.050 \scriptscriptstyle \pm \scriptstyle 3.4e\text{-}2$ & $0.870 \scriptscriptstyle \pm \scriptstyle 1.0e\text{-}3$ & $0.819 \scriptscriptstyle \pm \scriptstyle 2.0e\text{-}3$ & $0.728 \scriptscriptstyle \pm \scriptstyle 1.6e\text{-}3$ & $5.525 \scriptscriptstyle \pm \scriptstyle 3.5e\text{-}2$ & $0.924 \scriptscriptstyle \pm \scriptstyle 4.0e\text{-}4$ & $0.970 \scriptscriptstyle \pm \scriptstyle 9.5e\text{-}4$ & $0.746 \scriptscriptstyle \pm \scriptstyle 3.0e\text{-}4$ \\
MLP-PLRLR       & $0.676 \scriptscriptstyle \pm \scriptstyle 1.6e\text{-}2$ & $0.863 \scriptscriptstyle \pm \scriptstyle 3.1e\text{-}3$ & $0.456 \scriptscriptstyle \pm \scriptstyle 3.7e\text{-}3$ & $3.038 \scriptscriptstyle \pm \scriptstyle 2.3e\text{-}2$ & $0.871 \scriptscriptstyle \pm \scriptstyle 1.4e\text{-}3$ & $0.818 \scriptscriptstyle \pm \scriptstyle 1.7e\text{-}3$ & $0.725 \scriptscriptstyle \pm \scriptstyle 1.6e\text{-}3$ & $5.606 \scriptscriptstyle \pm \scriptstyle 8.9e\text{-}2$ & $0.924 \scriptscriptstyle \pm \scriptstyle 2.8e\text{-}4$ & $0.968 \scriptscriptstyle \pm \scriptstyle 2.0e\text{-}3$ & $0.744 \scriptscriptstyle \pm \scriptstyle 2.8e\text{-}4$ \\
MLP-AutoDis     & $0.649 \scriptscriptstyle \pm \scriptstyle 1.2e\text{-}2$ & $0.857 \scriptscriptstyle \pm \scriptstyle 3.2e\text{-}3$ & $0.474 \scriptscriptstyle \pm \scriptstyle 5.1e\text{-}3$ & $3.165 \scriptscriptstyle \pm \scriptstyle 1.8e\text{-}2$ & $0.859 \scriptscriptstyle \pm \scriptstyle 1.3e\text{-}3$ & $0.807 \scriptscriptstyle \pm \scriptstyle 2.4e\text{-}3$ & $0.725 \scriptscriptstyle \pm \scriptstyle 1.9e\text{-}3$ & $5.670 \scriptscriptstyle \pm \scriptstyle 6.1e\text{-}2$ & $0.924 \scriptscriptstyle \pm \scriptstyle 3.0e\text{-}4$ & $0.963 \scriptscriptstyle \pm \scriptstyle 8.7e\text{-}4$ & -- \\
MLP-DICE        & $0.610 \scriptscriptstyle \pm \scriptstyle 1.2e\text{-}2$ & $0.858 \scriptscriptstyle \pm \scriptstyle 2.9e\text{-}3$ & $0.491 \scriptscriptstyle \pm \scriptstyle 3.0e\text{-}3$ & $3.146 \scriptscriptstyle \pm \scriptstyle 3.5e\text{-}2$ & $0.860 \scriptscriptstyle \pm \scriptstyle 1.4e\text{-}3$ & $0.778 \scriptscriptstyle \pm \scriptstyle 4.9e\text{-}3$ & $0.720 \scriptscriptstyle \pm \scriptstyle 9.8e\text{-}4$ & $5.726 \scriptscriptstyle \pm \scriptstyle 3.6e\text{-}2$ & $0.920 \scriptscriptstyle \pm \scriptstyle 4.8e\text{-}4$ & $0.964 \scriptscriptstyle \pm \scriptstyle 1.1e\text{-}3$ & $0.748 \scriptscriptstyle \pm \scriptstyle 2.9e\text{-}4$ \\
\midrule
ResNet          & $0.655 \scriptscriptstyle \pm \scriptstyle 2.0e\text{-}2$ & $0.858 \scriptscriptstyle \pm \scriptstyle 3.1e\text{-}3$ & $0.490 \scriptscriptstyle \pm \scriptstyle 5.0e\text{-}3$ & $3.153 \scriptscriptstyle \pm \scriptstyle 3.6e\text{-}2$ & $0.855 \scriptscriptstyle \pm \scriptstyle 8.9e\text{-}4$ & $0.817 \scriptscriptstyle \pm \scriptstyle 3.4e\text{-}3$ & $0.729 \scriptscriptstyle \pm \scriptstyle 2.1e\text{-}3$ & $5.681 \scriptscriptstyle \pm \scriptstyle 5.3e\text{-}2$ & $0.916 \scriptscriptstyle \pm \scriptstyle 5.0e\text{-}4$ & $0.965 \scriptscriptstyle \pm \scriptstyle 8.3e\text{-}4$ & $0.747 \scriptscriptstyle \pm \scriptstyle 4.1e\text{-}4$ \\
ResNet-L        & $0.644 \scriptscriptstyle \pm \scriptstyle 1.9e\text{-}2$ & $0.859 \scriptscriptstyle \pm \scriptstyle 1.8e\text{-}3$ & $0.490 \scriptscriptstyle \pm \scriptstyle 6.6e\text{-}3$ & $3.126 \scriptscriptstyle \pm \scriptstyle 5.6e\text{-}2$ & $0.855 \scriptscriptstyle \pm \scriptstyle 1.4e\text{-}3$ & $0.813 \scriptscriptstyle \pm \scriptstyle 2.2e\text{-}3$ & $0.730 \scriptscriptstyle \pm \scriptstyle 9.7e\text{-}4$ & $5.758 \scriptscriptstyle \pm \scriptstyle 8.0e\text{-}2$ & $0.915 \scriptscriptstyle \pm \scriptstyle 4.3e\text{-}4$ & $0.964 \scriptscriptstyle \pm \scriptstyle 1.8e\text{-}3$ & $0.747 \scriptscriptstyle \pm \scriptstyle 4.7e\text{-}4$ \\
ResNet-LR       & $0.635 \scriptscriptstyle \pm \scriptstyle 2.3e\text{-}2$ & $0.861 \scriptscriptstyle \pm \scriptstyle 2.2e\text{-}3$ & $0.465 \scriptscriptstyle \pm \scriptstyle 3.5e\text{-}3$ & $3.096 \scriptscriptstyle \pm \scriptstyle 5.8e\text{-}2$ & $0.856 \scriptscriptstyle \pm \scriptstyle 1.6e\text{-}3$ & $0.815 \scriptscriptstyle \pm \scriptstyle 3.3e\text{-}3$ & $0.729 \scriptscriptstyle \pm \scriptstyle 1.3e\text{-}3$ & $5.574 \scriptscriptstyle \pm \scriptstyle 7.4e\text{-}2$ & $0.922 \scriptscriptstyle \pm \scriptstyle 4.4e\text{-}4$ & $0.967 \scriptscriptstyle \pm \scriptstyle 8.8e\text{-}4$ & $0.746 \scriptscriptstyle \pm \scriptstyle 4.4e\text{-}4$ \\
ResNet-Q        & $0.658 \scriptscriptstyle \pm \scriptstyle 8.0e\text{-}3$ & $0.858 \scriptscriptstyle \pm \scriptstyle 2.4e\text{-}3$ & $0.454 \scriptscriptstyle \pm \scriptstyle 3.6e\text{-}3$ & $3.251 \scriptscriptstyle \pm \scriptstyle 3.8e\text{-}2$ & $0.860 \scriptscriptstyle \pm \scriptstyle 1.3e\text{-}3$ & $0.811 \scriptscriptstyle \pm \scriptstyle 1.6e\text{-}3$ & $0.718 \scriptscriptstyle \pm \scriptstyle 1.0e\text{-}3$ & $5.828 \scriptscriptstyle \pm \scriptstyle 9.3e\text{-}2$ & $0.921 \scriptscriptstyle \pm \scriptstyle 9.1e\text{-}4$ & $0.970 \scriptscriptstyle \pm \scriptstyle 5.7e\text{-}4$ & $0.749 \scriptscriptstyle \pm \scriptstyle 2.9e\text{-}4$ \\
ResNet-Q-LR     & $0.650 \scriptscriptstyle \pm \scriptstyle 9.2e\text{-}3$ & $0.854 \scriptscriptstyle \pm \scriptstyle 4.2e\text{-}3$ & $0.446 \scriptscriptstyle \pm \scriptstyle 5.1e\text{-}3$ & $3.217 \scriptscriptstyle \pm \scriptstyle 5.2e\text{-}2$ & $0.865 \scriptscriptstyle \pm \scriptstyle 2.2e\text{-}3$ & $0.808 \scriptscriptstyle \pm \scriptstyle 2.6e\text{-}3$ & $0.722 \scriptscriptstyle \pm \scriptstyle 1.9e\text{-}3$ & $5.514 \scriptscriptstyle \pm \scriptstyle 6.0e\text{-}2$ & $0.922 \scriptscriptstyle \pm \scriptstyle 5.9e\text{-}4$ & $0.972 \scriptscriptstyle \pm \scriptstyle 3.7e\text{-}4$ & $0.748 \scriptscriptstyle \pm \scriptstyle 5.0e\text{-}4$ \\
ResNet-T        & $0.657 \scriptscriptstyle \pm \scriptstyle 9.0e\text{-}3$ & $0.859 \scriptscriptstyle \pm \scriptstyle 2.9e\text{-}3$ & $0.441 \scriptscriptstyle \pm \scriptstyle 3.2e\text{-}3$ & $3.151 \scriptscriptstyle \pm \scriptstyle 5.9e\text{-}2$ & $0.866 \scriptscriptstyle \pm \scriptstyle 1.8e\text{-}3$ & $0.817 \scriptscriptstyle \pm \scriptstyle 1.7e\text{-}3$ & $0.724 \scriptscriptstyle \pm \scriptstyle 2.0e\text{-}3$ & $5.781 \scriptscriptstyle \pm \scriptstyle 4.1e\text{-}2$ & $0.923 \scriptscriptstyle \pm \scriptstyle 6.0e\text{-}4$ & $0.970 \scriptscriptstyle \pm \scriptstyle 1.1e\text{-}3$ & $0.749 \scriptscriptstyle \pm \scriptstyle 7.8e\text{-}4$ \\
ResNet-T-LR     & $0.650 \scriptscriptstyle \pm \scriptstyle 1.2e\text{-}2$ & $0.861 \scriptscriptstyle \pm \scriptstyle 2.0e\text{-}3$ & $0.438 \scriptscriptstyle \pm \scriptstyle 2.9e\text{-}3$ & $3.163 \scriptscriptstyle \pm \scriptstyle 6.1e\text{-}2$ & $0.870 \scriptscriptstyle \pm \scriptstyle 1.5e\text{-}3$ & $0.813 \scriptscriptstyle \pm \scriptstyle 2.5e\text{-}3$ & $0.725 \scriptscriptstyle \pm \scriptstyle 1.6e\text{-}3$ & $5.687 \scriptscriptstyle \pm \scriptstyle 5.9e\text{-}2$ & $0.922 \scriptscriptstyle \pm \scriptstyle 8.1e\text{-}4$ & $0.972 \scriptscriptstyle \pm \scriptstyle 3.7e\text{-}4$ & $0.748 \scriptscriptstyle \pm \scriptstyle 6.1e\text{-}4$ \\
ResNet-P        & $0.630 \scriptscriptstyle \pm \scriptstyle 1.8e\text{-}2$ & $0.858 \scriptscriptstyle \pm \scriptstyle 3.1e\text{-}3$ & $0.471 \scriptscriptstyle \pm \scriptstyle 6.5e\text{-}3$ & $3.147 \scriptscriptstyle \pm \scriptstyle 2.9e\text{-}2$ & $0.866 \scriptscriptstyle \pm \scriptstyle 1.7e\text{-}3$ & $0.812 \scriptscriptstyle \pm \scriptstyle 1.6e\text{-}3$ & $0.729 \scriptscriptstyle \pm \scriptstyle 7.0e\text{-}4$ & $5.566 \scriptscriptstyle \pm \scriptstyle 7.5e\text{-}2$ & $0.922 \scriptscriptstyle \pm \scriptstyle 6.7e\text{-}4$ & $0.968 \scriptscriptstyle \pm \scriptstyle 7.7e\text{-}4$ & $0.747 \scriptscriptstyle \pm \scriptstyle 6.3e\text{-}4$ \\
ResNet-PLR      & $0.651 \scriptscriptstyle \pm \scriptstyle 1.3e\text{-}2$ & $0.859 \scriptscriptstyle \pm \scriptstyle 3.7e\text{-}3$ & $0.461 \scriptscriptstyle \pm \scriptstyle 4.2e\text{-}3$ & $3.188 \scriptscriptstyle \pm \scriptstyle 7.3e\text{-}2$ & $0.869 \scriptscriptstyle \pm \scriptstyle 1.7e\text{-}3$ & $0.816 \scriptscriptstyle \pm \scriptstyle 2.5e\text{-}3$ & $0.728 \scriptscriptstyle \pm \scriptstyle 1.8e\text{-}3$ & $5.582 \scriptscriptstyle \pm \scriptstyle 4.9e\text{-}2$ & $0.923 \scriptscriptstyle \pm \scriptstyle 5.9e\text{-}4$ & $0.972 \scriptscriptstyle \pm \scriptstyle 5.1e\text{-}4$ & $0.747 \scriptscriptstyle \pm \scriptstyle 6.4e\text{-}4$ \\
\midrule
Transformer-L   & $0.632 \scriptscriptstyle \pm \scriptstyle 2.0e\text{-}2$ & $0.860 \scriptscriptstyle \pm \scriptstyle 3.0e\text{-}3$ & $0.465 \scriptscriptstyle \pm \scriptstyle 4.8e\text{-}3$ & $3.239 \scriptscriptstyle \pm \scriptstyle 3.2e\text{-}2$ & $0.858 \scriptscriptstyle \pm \scriptstyle 1.3e\text{-}3$ & $0.817 \scriptscriptstyle \pm \scriptstyle 2.3e\text{-}3$ & $0.725 \scriptscriptstyle \pm \scriptstyle 3.2e\text{-}3$ & $5.602 \scriptscriptstyle \pm \scriptstyle 4.8e\text{-}2$ & $0.924 \scriptscriptstyle \pm \scriptstyle 4.4e\text{-}4$ & $0.971 \scriptscriptstyle \pm \scriptstyle 6.8e\text{-}4$ & $0.746 \scriptscriptstyle \pm \scriptstyle 5.7e\text{-}4$ \\
Transformer-LR  & $0.614 \scriptscriptstyle \pm \scriptstyle 4.5e\text{-}2$ & $0.860 \scriptscriptstyle \pm \scriptstyle 2.2e\text{-}3$ & $0.456 \scriptscriptstyle \pm \scriptstyle 3.7e\text{-}3$ & $3.261 \scriptscriptstyle \pm \scriptstyle 5.6e\text{-}2$ & $0.858 \scriptscriptstyle \pm \scriptstyle 1.6e\text{-}3$ & $0.817 \scriptscriptstyle \pm \scriptstyle 2.2e\text{-}3$ & $0.729 \scriptscriptstyle \pm \scriptstyle 1.5e\text{-}3$ & $5.644 \scriptscriptstyle \pm \scriptstyle 5.5e\text{-}2$ & $0.924 \scriptscriptstyle \pm \scriptstyle 3.9e\text{-}4$ & $0.971 \scriptscriptstyle \pm \scriptstyle 7.6e\text{-}4$ & $0.746 \scriptscriptstyle \pm \scriptstyle 5.8e\text{-}4$ \\
Transformer-Q-L & $0.659 \scriptscriptstyle \pm \scriptstyle 8.7e\text{-}3$ & $0.856 \scriptscriptstyle \pm \scriptstyle 5.9e\text{-}3$ & $0.451 \scriptscriptstyle \pm \scriptstyle 5.4e\text{-}3$ & $3.319 \scriptscriptstyle \pm \scriptstyle 4.2e\text{-}2$ & $0.867 \scriptscriptstyle \pm \scriptstyle 1.6e\text{-}3$ & $0.812 \scriptscriptstyle \pm \scriptstyle 2.6e\text{-}3$ & $0.729 \scriptscriptstyle \pm \scriptstyle 2.9e\text{-}3$ & $5.741 \scriptscriptstyle \pm \scriptstyle 4.5e\text{-}2$ & $0.924 \scriptscriptstyle \pm \scriptstyle 3.8e\text{-}4$ & $0.973 \scriptscriptstyle \pm \scriptstyle 6.1e\text{-}4$ & $0.747 \scriptscriptstyle \pm \scriptstyle 7.9e\text{-}4$ \\
Transformer-Q-LR & $0.659 \scriptscriptstyle \pm \scriptstyle 1.2e\text{-}2$ & $0.857 \scriptscriptstyle \pm \scriptstyle 2.0e\text{-}3$ & $0.448 \scriptscriptstyle \pm \scriptstyle 6.1e\text{-}3$ & $3.270 \scriptscriptstyle \pm \scriptstyle 4.6e\text{-}2$ & $0.867 \scriptscriptstyle \pm \scriptstyle 1.1e\text{-}3$ & $0.812 \scriptscriptstyle \pm \scriptstyle 2.5e\text{-}3$ & $0.723 \scriptscriptstyle \pm \scriptstyle 3.3e\text{-}3$ & $5.683 \scriptscriptstyle \pm \scriptstyle 4.8e\text{-}2$ & $0.923 \scriptscriptstyle \pm \scriptstyle 5.8e\text{-}4$ & $0.972 \scriptscriptstyle \pm \scriptstyle 4.2e\text{-}4$ & $0.748 \scriptscriptstyle \pm \scriptstyle 7.7e\text{-}4$ \\
Transformer-T-L & $0.663 \scriptscriptstyle \pm \scriptstyle 7.4e\text{-}3$ & $0.861 \scriptscriptstyle \pm \scriptstyle 1.4e\text{-}3$ & $0.454 \scriptscriptstyle \pm \scriptstyle 4.7e\text{-}3$ & $3.197 \scriptscriptstyle \pm \scriptstyle 2.9e\text{-}2$ & $0.871 \scriptscriptstyle \pm \scriptstyle 1.4e\text{-}3$ & $0.817 \scriptscriptstyle \pm \scriptstyle 2.6e\text{-}3$ & $0.726 \scriptscriptstyle \pm \scriptstyle 1.7e\text{-}3$ & $5.803 \scriptscriptstyle \pm \scriptstyle 6.5e\text{-}2$ & $0.924 \scriptscriptstyle \pm \scriptstyle 3.3e\text{-}4$ & $0.974 \scriptscriptstyle \pm \scriptstyle 4.5e\text{-}4$ & $0.747 \scriptscriptstyle \pm \scriptstyle 6.5e\text{-}4$ \\
Transformer-T-LR & $0.665 \scriptscriptstyle \pm \scriptstyle 6.6e\text{-}3$ & $0.860 \scriptscriptstyle \pm \scriptstyle 3.4e\text{-}3$ & $0.442 \scriptscriptstyle \pm \scriptstyle 5.3e\text{-}3$ & $3.219 \scriptscriptstyle \pm \scriptstyle 3.2e\text{-}2$ & $0.870 \scriptscriptstyle \pm \scriptstyle 1.5e\text{-}3$ & $0.818 \scriptscriptstyle \pm \scriptstyle 2.6e\text{-}3$ & $0.729 \scriptscriptstyle \pm \scriptstyle 1.4e\text{-}3$ & $5.699 \scriptscriptstyle \pm \scriptstyle 6.7e\text{-}2$ & $0.924 \scriptscriptstyle \pm \scriptstyle 4.4e\text{-}4$ & $0.973 \scriptscriptstyle \pm \scriptstyle 5.6e\text{-}4$ & $0.747 \scriptscriptstyle \pm \scriptstyle 8.4e\text{-}4$ \\
Transformer-PLR & $0.646 \scriptscriptstyle \pm \scriptstyle 2.0e\text{-}2$ & $0.863 \scriptscriptstyle \pm \scriptstyle 2.7e\text{-}3$ & $0.464 \scriptscriptstyle \pm \scriptstyle 2.8e\text{-}3$ & $3.162 \scriptscriptstyle \pm \scriptstyle 4.2e\text{-}2$ & $0.870 \scriptscriptstyle \pm \scriptstyle 1.5e\text{-}3$ & $0.814 \scriptscriptstyle \pm \scriptstyle 2.1e\text{-}3$ & $0.730 \scriptscriptstyle \pm \scriptstyle 1.9e\text{-}3$ & $5.760 \scriptscriptstyle \pm \scriptstyle 1.1e\text{-}1$ & $0.924 \scriptscriptstyle \pm \scriptstyle 5.2e\text{-}4$ & $0.972 \scriptscriptstyle \pm \scriptstyle 1.1e\text{-}3$ & $0.746 \scriptscriptstyle \pm \scriptstyle 5.9e\text{-}4$ \\
\bottomrule
\end{tabular}}
\end{sidewaystable}    

\begin{sidewaystable}[t!]
    \setlength\tabcolsep{2.2pt}
    \centering
    \caption{Extended results for ensembles}
    \label{tab:A_ensembles}
    \vspace{1em}
    {\scriptsize \begin{tabular}{lccccccccccc}
\toprule
{} & GE \textuparrow & CH \textuparrow & CA \textdownarrow & HO \textdownarrow & AD \textuparrow & OT \textuparrow & HI \textuparrow & FB \textdownarrow & SA \textuparrow & CO \textuparrow & MI \textdownarrow \\
\midrule
CatBoost        & $0.692 \scriptscriptstyle \pm \scriptstyle 1.9e\text{-}3$ & $0.861 \scriptscriptstyle \pm \scriptstyle 2.4e\text{-}4$ & $0.430 \scriptscriptstyle \pm \scriptstyle 1.1e\text{-}3$ & $3.093 \scriptscriptstyle \pm \scriptstyle 5.1e\text{-}3$ & $0.873 \scriptscriptstyle \pm \scriptstyle 5.1e\text{-}4$ & $0.825 \scriptscriptstyle \pm \scriptstyle 4.7e\text{-}4$ & $0.727 \scriptscriptstyle \pm \scriptstyle 3.6e\text{-}4$ & $5.226 \scriptscriptstyle \pm \scriptstyle 1.3e\text{-}2$ & $0.924 \scriptscriptstyle \pm \scriptstyle 1.0e\text{-}4$ & $0.967 \scriptscriptstyle \pm \scriptstyle 1.4e\text{-}4$ & $0.741 \scriptscriptstyle \pm \scriptstyle 1.4e\text{-}4$ \\
XGBoost         & $0.683 \scriptscriptstyle \pm \scriptstyle 1.3e\text{-}3$ & $0.859 \scriptscriptstyle \pm \scriptstyle 2.4e\text{-}4$ & $0.434 \scriptscriptstyle \pm \scriptstyle 7.1e\text{-}4$ & $3.152 \scriptscriptstyle \pm \scriptstyle 1.2e\text{-}3$ & $0.875 \scriptscriptstyle \pm \scriptstyle 5.5e\text{-}4$ & $0.827 \scriptscriptstyle \pm \scriptstyle 8.4e\text{-}4$ & $0.726 \scriptscriptstyle \pm \scriptstyle 8.1e\text{-}4$ & $5.338 \scriptscriptstyle \pm \scriptstyle 1.9e\text{-}2$ & $0.919 \scriptscriptstyle \pm \scriptstyle 4.8e\text{-}4$ & $0.969 \scriptscriptstyle \pm \scriptstyle 8.8e\text{-}5$ & $0.742 \scriptscriptstyle \pm \scriptstyle 5.3e\text{-}5$ \\
\midrule
MLP             & $0.665 \scriptscriptstyle \pm \scriptstyle 2.7e\text{-}3$ & $0.856 \scriptscriptstyle \pm \scriptstyle 1.2e\text{-}3$ & $0.486 \scriptscriptstyle \pm \scriptstyle 7.8e\text{-}4$ & $3.109 \scriptscriptstyle \pm \scriptstyle 1.0e\text{-}2$ & $0.856 \scriptscriptstyle \pm \scriptstyle 4.6e\text{-}4$ & $0.822 \scriptscriptstyle \pm \scriptstyle 8.0e\text{-}4$ & $0.727 \scriptscriptstyle \pm \scriptstyle 1.7e\text{-}3$ & $5.616 \scriptscriptstyle \pm \scriptstyle 7.6e\text{-}3$ & $0.913 \scriptscriptstyle \pm \scriptstyle 8.2e\text{-}5$ & $0.968 \scriptscriptstyle \pm \scriptstyle 4.8e\text{-}4$ & $0.746 \scriptscriptstyle \pm \scriptstyle 1.1e\text{-}4$ \\
MLP-L           & $0.670 \scriptscriptstyle \pm \scriptstyle 3.2e\text{-}3$ & $0.862 \scriptscriptstyle \pm \scriptstyle 1.5e\text{-}3$ & $0.471 \scriptscriptstyle \pm \scriptstyle 4.7e\text{-}4$ & $3.021 \scriptscriptstyle \pm \scriptstyle 1.1e\text{-}2$ & $0.857 \scriptscriptstyle \pm \scriptstyle 5.9e\text{-}4$ & $0.824 \scriptscriptstyle \pm \scriptstyle 1.1e\text{-}3$ & $0.728 \scriptscriptstyle \pm \scriptstyle 2.7e\text{-}4$ & $5.508 \scriptscriptstyle \pm \scriptstyle 2.1e\text{-}2$ & $0.916 \scriptscriptstyle \pm \scriptstyle 1.5e\text{-}4$ & $0.971 \scriptscriptstyle \pm \scriptstyle 6.8e\text{-}5$ & $0.746 \scriptscriptstyle \pm \scriptstyle 2.3e\text{-}4$ \\
MLP-LR          & $0.679 \scriptscriptstyle \pm \scriptstyle 4.9e\text{-}3$ & $0.861 \scriptscriptstyle \pm \scriptstyle 9.4e\text{-}4$ & $0.463 \scriptscriptstyle \pm \scriptstyle 1.9e\text{-}3$ & $3.012 \scriptscriptstyle \pm \scriptstyle 1.8e\text{-}3$ & $0.859 \scriptscriptstyle \pm \scriptstyle 8.0e\text{-}4$ & $0.826 \scriptscriptstyle \pm \scriptstyle 1.6e\text{-}3$ & $0.731 \scriptscriptstyle \pm \scriptstyle 1.1e\text{-}3$ & $5.477 \scriptscriptstyle \pm \scriptstyle 3.6e\text{-}2$ & $0.924 \scriptscriptstyle \pm \scriptstyle 7.1e\text{-}5$ & $0.972 \scriptscriptstyle \pm \scriptstyle 7.6e\text{-}5$ & $0.744 \scriptscriptstyle \pm \scriptstyle 1.6e\text{-}4$ \\
MLP-LRLR        & $0.676 \scriptscriptstyle \pm \scriptstyle 4.8e\text{-}3$ & $0.863 \scriptscriptstyle \pm \scriptstyle 1.4e\text{-}3$ & $0.453 \scriptscriptstyle \pm \scriptstyle 1.1e\text{-}3$ & $3.017 \scriptscriptstyle \pm \scriptstyle 1.1e\text{-}2$ & $0.858 \scriptscriptstyle \pm \scriptstyle 1.6e\text{-}4$ & $0.828 \scriptscriptstyle \pm \scriptstyle 1.3e\text{-}3$ & $0.725 \scriptscriptstyle \pm \scriptstyle 5.9e\text{-}4$ & $5.427 \scriptscriptstyle \pm \scriptstyle 2.1e\text{-}2$ & $0.924 \scriptscriptstyle \pm \scriptstyle 1.2e\text{-}4$ & $0.973 \scriptscriptstyle \pm \scriptstyle 1.8e\text{-}4$ & $0.744 \scriptscriptstyle \pm \scriptstyle 1.8e\text{-}4$ \\
MLP-Q           & $0.677 \scriptscriptstyle \pm \scriptstyle 4.8e\text{-}3$ & $0.856 \scriptscriptstyle \pm \scriptstyle 1.4e\text{-}3$ & $0.458 \scriptscriptstyle \pm \scriptstyle 1.7e\text{-}4$ & $3.080 \scriptscriptstyle \pm \scriptstyle 1.5e\text{-}2$ & $0.862 \scriptscriptstyle \pm \scriptstyle 4.0e\text{-}4$ & $0.822 \scriptscriptstyle \pm \scriptstyle 1.7e\text{-}3$ & $0.723 \scriptscriptstyle \pm \scriptstyle 5.6e\text{-}4$ & $5.706 \scriptscriptstyle \pm \scriptstyle 1.9e\text{-}2$ & $0.922 \scriptscriptstyle \pm \scriptstyle 1.7e\text{-}4$ & $0.973 \scriptscriptstyle \pm \scriptstyle 2.1e\text{-}4$ & $0.748 \scriptscriptstyle \pm \scriptstyle 2.2e\text{-}4$ \\
MLP-Q-LR        & $0.682 \scriptscriptstyle \pm \scriptstyle 3.9e\text{-}3$ & $0.859 \scriptscriptstyle \pm \scriptstyle 4.7e\text{-}4$ & $0.433 \scriptscriptstyle \pm \scriptstyle 1.9e\text{-}3$ & $3.080 \scriptscriptstyle \pm \scriptstyle 9.7e\text{-}3$ & $0.867 \scriptscriptstyle \pm \scriptstyle 4.2e\text{-}4$ & $0.818 \scriptscriptstyle \pm \scriptstyle 1.4e\text{-}3$ & $0.724 \scriptscriptstyle \pm \scriptstyle 3.2e\text{-}4$ & $5.144 \scriptscriptstyle \pm \scriptstyle 1.4e\text{-}2$ & $0.924 \scriptscriptstyle \pm \scriptstyle 3.7e\text{-}4$ & $0.974 \scriptscriptstyle \pm \scriptstyle 1.5e\text{-}4$ & $0.745 \scriptscriptstyle \pm \scriptstyle 2.8e\text{-}4$ \\
MLP-Q-LRLR      & $0.674 \scriptscriptstyle \pm \scriptstyle 2.9e\text{-}3$ & $0.862 \scriptscriptstyle \pm \scriptstyle 1.5e\text{-}3$ & $0.438 \scriptscriptstyle \pm \scriptstyle 2.1e\text{-}3$ & $3.066 \scriptscriptstyle \pm \scriptstyle 9.9e\text{-}3$ & $0.870 \scriptscriptstyle \pm \scriptstyle 4.1e\text{-}4$ & $0.817 \scriptscriptstyle \pm \scriptstyle 2.4e\text{-}3$ & $0.727 \scriptscriptstyle \pm \scriptstyle 2.1e\text{-}4$ & $5.268 \scriptscriptstyle \pm \scriptstyle 7.5e\text{-}2$ & $0.924 \scriptscriptstyle \pm \scriptstyle 3.1e\text{-}5$ & $0.973 \scriptscriptstyle \pm \scriptstyle 2.8e\text{-}4$ & $0.745 \scriptscriptstyle \pm \scriptstyle 1.7e\text{-}4$ \\
MLP-T           & $0.669 \scriptscriptstyle \pm \scriptstyle 4.3e\text{-}3$ & $0.861 \scriptscriptstyle \pm \scriptstyle 1.0e\text{-}3$ & $0.439 \scriptscriptstyle \pm \scriptstyle 2.1e\text{-}4$ & $3.058 \scriptscriptstyle \pm \scriptstyle 1.4e\text{-}2$ & $0.865 \scriptscriptstyle \pm \scriptstyle 5.3e\text{-}4$ & $0.822 \scriptscriptstyle \pm \scriptstyle 6.3e\text{-}4$ & $0.724 \scriptscriptstyle \pm \scriptstyle 7.2e\text{-}4$ & $5.507 \scriptscriptstyle \pm \scriptstyle 2.0e\text{-}2$ & $0.923 \scriptscriptstyle \pm \scriptstyle 8.5e\text{-}5$ & $0.972 \scriptscriptstyle \pm \scriptstyle 2.7e\text{-}4$ & $0.747 \scriptscriptstyle \pm \scriptstyle 4.1e\text{-}5$ \\
MLP-T-LR        & $0.673 \scriptscriptstyle \pm \scriptstyle 8.3e\text{-}4$ & $0.861 \scriptscriptstyle \pm \scriptstyle 8.5e\text{-}4$ & $0.435 \scriptscriptstyle \pm \scriptstyle 1.1e\text{-}3$ & $3.099 \scriptscriptstyle \pm \scriptstyle 2.4e\text{-}2$ & $0.870 \scriptscriptstyle \pm \scriptstyle 6.6e\text{-}4$ & $0.821 \scriptscriptstyle \pm \scriptstyle 2.6e\text{-}4$ & $0.727 \scriptscriptstyle \pm \scriptstyle 7.2e\text{-}4$ & $5.409 \scriptscriptstyle \pm \scriptstyle 6.2e\text{-}3$ & $0.924 \scriptscriptstyle \pm \scriptstyle 1.3e\text{-}4$ & $0.973 \scriptscriptstyle \pm \scriptstyle 1.3e\text{-}4$ & $0.746 \scriptscriptstyle \pm \scriptstyle 1.6e\text{-}4$ \\
MLP-T-LRLR      & $0.670 \scriptscriptstyle \pm \scriptstyle 4.1e\text{-}4$ & $0.860 \scriptscriptstyle \pm \scriptstyle 2.5e\text{-}3$ & $0.431 \scriptscriptstyle \pm \scriptstyle 6.0e\text{-}4$ & $3.056 \scriptscriptstyle \pm \scriptstyle 2.2e\text{-}2$ & $0.870 \scriptscriptstyle \pm \scriptstyle 2.6e\text{-}4$ & $0.826 \scriptscriptstyle \pm \scriptstyle 5.0e\text{-}4$ & $0.725 \scriptscriptstyle \pm \scriptstyle 7.4e\text{-}4$ & $5.440 \scriptscriptstyle \pm \scriptstyle 1.8e\text{-}3$ & $0.925 \scriptscriptstyle \pm \scriptstyle 6.1e\text{-}5$ & $0.973 \scriptscriptstyle \pm \scriptstyle 2.2e\text{-}4$ & $0.745 \scriptscriptstyle \pm \scriptstyle 4.7e\text{-}4$ \\
MLP-P           & $0.661 \scriptscriptstyle \pm \scriptstyle 6.0e\text{-}3$ & $0.861 \scriptscriptstyle \pm \scriptstyle 6.2e\text{-}4$ & $0.473 \scriptscriptstyle \pm \scriptstyle 1.1e\text{-}3$ & $3.042 \scriptscriptstyle \pm \scriptstyle 1.0e\text{-}2$ & $0.871 \scriptscriptstyle \pm \scriptstyle 1.1e\text{-}3$ & $0.812 \scriptscriptstyle \pm \scriptstyle 1.7e\text{-}3$ & $0.725 \scriptscriptstyle \pm \scriptstyle 6.2e\text{-}4$ & $5.508 \scriptscriptstyle \pm \scriptstyle 3.1e\text{-}2$ & $0.924 \scriptscriptstyle \pm \scriptstyle 5.4e\text{-}5$ & $0.973 \scriptscriptstyle \pm \scriptstyle 3.0e\text{-}4$ & $0.745 \scriptscriptstyle \pm \scriptstyle 2.1e\text{-}4$ \\
MLP-PL          & $0.671 \scriptscriptstyle \pm \scriptstyle 6.2e\text{-}3$ & $0.860 \scriptscriptstyle \pm \scriptstyle 1.2e\text{-}3$ & $0.456 \scriptscriptstyle \pm \scriptstyle 1.3e\text{-}3$ & $3.065 \scriptscriptstyle \pm \scriptstyle 8.1e\text{-}3$ & $0.872 \scriptscriptstyle \pm \scriptstyle 6.3e\text{-}4$ & $0.825 \scriptscriptstyle \pm \scriptstyle 4.1e\text{-}4$ & $0.730 \scriptscriptstyle \pm \scriptstyle 3.5e\text{-}4$ & $5.216 \scriptscriptstyle \pm \scriptstyle 2.0e\text{-}2$ & $0.924 \scriptscriptstyle \pm \scriptstyle 1.2e\text{-}4$ & $0.974 \scriptscriptstyle \pm \scriptstyle 1.9e\text{-}4$ & $0.744 \scriptscriptstyle \pm \scriptstyle 2.1e\text{-}4$ \\
MLP-PLR         & $0.700 \scriptscriptstyle \pm \scriptstyle 2.1e\text{-}3$ & $0.858 \scriptscriptstyle \pm \scriptstyle 1.6e\text{-}3$ & $0.453 \scriptscriptstyle \pm \scriptstyle 5.8e\text{-}4$ & $2.975 \scriptscriptstyle \pm \scriptstyle 6.6e\text{-}3$ & $0.874 \scriptscriptstyle \pm \scriptstyle 9.0e\text{-}4$ & $0.830 \scriptscriptstyle \pm \scriptstyle 2.4e\text{-}3$ & $0.734 \scriptscriptstyle \pm \scriptstyle 3.5e\text{-}4$ & $5.388 \scriptscriptstyle \pm \scriptstyle 1.6e\text{-}2$ & $0.924 \scriptscriptstyle \pm \scriptstyle 5.4e\text{-}5$ & $0.975 \scriptscriptstyle \pm \scriptstyle 4.8e\text{-}4$ & $0.743 \scriptscriptstyle \pm \scriptstyle 1.0e\text{-}4$ \\
MLP-PLRLR       & $0.699 \scriptscriptstyle \pm \scriptstyle 9.3e\text{-}3$ & $0.867 \scriptscriptstyle \pm \scriptstyle 1.8e\text{-}3$ & $0.448 \scriptscriptstyle \pm \scriptstyle 8.3e\text{-}4$ & $2.993 \scriptscriptstyle \pm \scriptstyle 6.5e\text{-}3$ & $0.873 \scriptscriptstyle \pm \scriptstyle 4.1e\text{-}4$ & $0.823 \scriptscriptstyle \pm \scriptstyle 8.3e\text{-}4$ & $0.729 \scriptscriptstyle \pm \scriptstyle 9.1e\text{-}4$ & $5.346 \scriptscriptstyle \pm \scriptstyle 4.8e\text{-}2$ & $0.924 \scriptscriptstyle \pm \scriptstyle 2.6e\text{-}4$ & $0.972 \scriptscriptstyle \pm \scriptstyle 8.2e\text{-}4$ & $0.743 \scriptscriptstyle \pm \scriptstyle 9.9e\text{-}5$ \\
MLP-AutoDis     & $0.676 \scriptscriptstyle \pm \scriptstyle 7.6e\text{-}3$ & $0.860 \scriptscriptstyle \pm \scriptstyle 1.7e\text{-}3$ & $0.464 \scriptscriptstyle \pm \scriptstyle 1.6e\text{-}3$ & $3.132 \scriptscriptstyle \pm \scriptstyle 5.7e\text{-}3$ & $0.860 \scriptscriptstyle \pm \scriptstyle 2.8e\text{-}4$ & $0.817 \scriptscriptstyle \pm \scriptstyle 2.1e\text{-}3$ & $0.730 \scriptscriptstyle \pm \scriptstyle 2.5e\text{-}4$ & $5.580 \scriptscriptstyle \pm \scriptstyle 2.2e\text{-}2$ & $0.924 \scriptscriptstyle \pm \scriptstyle 1.1e\text{-}4$ & $0.970 \scriptscriptstyle \pm \scriptstyle 3.2e\text{-}4$ & -- \\
MLP-DICE        & $0.636 \scriptscriptstyle \pm \scriptstyle 2.6e\text{-}3$ & $0.859 \scriptscriptstyle \pm \scriptstyle 2.0e\text{-}3$ & $0.486 \scriptscriptstyle \pm \scriptstyle 1.4e\text{-}3$ & $3.092 \scriptscriptstyle \pm \scriptstyle 1.3e\text{-}2$ & $0.862 \scriptscriptstyle \pm \scriptstyle 4.8e\text{-}4$ & $0.784 \scriptscriptstyle \pm \scriptstyle 2.3e\text{-}3$ & $0.723 \scriptscriptstyle \pm \scriptstyle 6.1e\text{-}4$ & $5.615 \scriptscriptstyle \pm \scriptstyle 8.9e\text{-}3$ & $0.920 \scriptscriptstyle \pm \scriptstyle 2.0e\text{-}4$ & $0.969 \scriptscriptstyle \pm \scriptstyle 1.4e\text{-}4$ & $0.746 \scriptscriptstyle \pm \scriptstyle 2.0e\text{-}4$ \\
\midrule
ResNet          & $0.690 \scriptscriptstyle \pm \scriptstyle 5.9e\text{-}3$ & $0.861 \scriptscriptstyle \pm \scriptstyle 1.6e\text{-}3$ & $0.483 \scriptscriptstyle \pm \scriptstyle 1.8e\text{-}3$ & $3.081 \scriptscriptstyle \pm \scriptstyle 7.8e\text{-}3$ & $0.856 \scriptscriptstyle \pm \scriptstyle 3.4e\text{-}4$ & $0.821 \scriptscriptstyle \pm \scriptstyle 1.8e\text{-}3$ & $0.734 \scriptscriptstyle \pm \scriptstyle 1.1e\text{-}3$ & $5.482 \scriptscriptstyle \pm \scriptstyle 1.1e\text{-}2$ & $0.918 \scriptscriptstyle \pm \scriptstyle 5.3e\text{-}4$ & $0.968 \scriptscriptstyle \pm \scriptstyle 4.4e\text{-}4$ & $0.745 \scriptscriptstyle \pm \scriptstyle 6.5e\text{-}5$ \\
ResNet-L        & $0.674 \scriptscriptstyle \pm \scriptstyle 5.2e\text{-}3$ & $0.859 \scriptscriptstyle \pm \scriptstyle 6.2e\text{-}4$ & $0.481 \scriptscriptstyle \pm \scriptstyle 2.5e\text{-}3$ & $3.025 \scriptscriptstyle \pm \scriptstyle 1.8e\text{-}2$ & $0.857 \scriptscriptstyle \pm \scriptstyle 2.9e\text{-}4$ & $0.819 \scriptscriptstyle \pm \scriptstyle 1.3e\text{-}3$ & $0.735 \scriptscriptstyle \pm \scriptstyle 5.2e\text{-}4$ & $5.522 \scriptscriptstyle \pm \scriptstyle 2.4e\text{-}2$ & $0.917 \scriptscriptstyle \pm \scriptstyle 2.2e\text{-}4$ & $0.966 \scriptscriptstyle \pm \scriptstyle 5.1e\text{-}4$ & $0.744 \scriptscriptstyle \pm \scriptstyle 3.0e\text{-}4$ \\
ResNet-LR       & $0.672 \scriptscriptstyle \pm \scriptstyle 6.0e\text{-}3$ & $0.862 \scriptscriptstyle \pm \scriptstyle 1.7e\text{-}3$ & $0.450 \scriptscriptstyle \pm \scriptstyle 2.2e\text{-}3$ & $2.992 \scriptscriptstyle \pm \scriptstyle 2.4e\text{-}2$ & $0.859 \scriptscriptstyle \pm \scriptstyle 4.7e\text{-}4$ & $0.822 \scriptscriptstyle \pm \scriptstyle 9.2e\text{-}4$ & $0.733 \scriptscriptstyle \pm \scriptstyle 4.2e\text{-}5$ & $5.415 \scriptscriptstyle \pm \scriptstyle 9.5e\text{-}5$ & $0.923 \scriptscriptstyle \pm \scriptstyle 7.7e\text{-}5$ & $0.971 \scriptscriptstyle \pm \scriptstyle 1.5e\text{-}4$ & $0.743 \scriptscriptstyle \pm \scriptstyle 2.1e\text{-}4$ \\
ResNet-Q        & $0.671 \scriptscriptstyle \pm \scriptstyle 1.7e\text{-}3$ & $0.862 \scriptscriptstyle \pm \scriptstyle 8.2e\text{-}4$ & $0.442 \scriptscriptstyle \pm \scriptstyle 8.0e\text{-}4$ & $3.128 \scriptscriptstyle \pm \scriptstyle 9.0e\text{-}3$ & $0.862 \scriptscriptstyle \pm \scriptstyle 5.8e\text{-}4$ & $0.816 \scriptscriptstyle \pm \scriptstyle 9.4e\text{-}4$ & $0.722 \scriptscriptstyle \pm \scriptstyle 7.1e\text{-}4$ & $5.402 \scriptscriptstyle \pm \scriptstyle 3.3e\text{-}2$ & $0.923 \scriptscriptstyle \pm \scriptstyle 4.6e\text{-}4$ & $0.974 \scriptscriptstyle \pm \scriptstyle 6.3e\text{-}5$ & $0.746 \scriptscriptstyle \pm \scriptstyle 2.4e\text{-}4$ \\
ResNet-Q-LR     & $0.674 \scriptscriptstyle \pm \scriptstyle 2.5e\text{-}3$ & $0.859 \scriptscriptstyle \pm \scriptstyle 1.8e\text{-}3$ & $0.427 \scriptscriptstyle \pm \scriptstyle 2.3e\text{-}3$ & $3.066 \scriptscriptstyle \pm \scriptstyle 2.2e\text{-}2$ & $0.868 \scriptscriptstyle \pm \scriptstyle 1.1e\text{-}3$ & $0.815 \scriptscriptstyle \pm \scriptstyle 7.1e\text{-}4$ & $0.729 \scriptscriptstyle \pm \scriptstyle 1.6e\text{-}3$ & $5.309 \scriptscriptstyle \pm \scriptstyle 4.9e\text{-}2$ & $0.923 \scriptscriptstyle \pm \scriptstyle 3.9e\text{-}4$ & $0.976 \scriptscriptstyle \pm \scriptstyle 1.2e\text{-}4$ & $0.746 \scriptscriptstyle \pm \scriptstyle 1.8e\text{-}4$ \\
ResNet-T        & $0.681 \scriptscriptstyle \pm \scriptstyle 1.3e\text{-}3$ & $0.861 \scriptscriptstyle \pm \scriptstyle 2.1e\text{-}3$ & $0.428 \scriptscriptstyle \pm \scriptstyle 8.0e\text{-}4$ & $3.064 \scriptscriptstyle \pm \scriptstyle 3.6e\text{-}2$ & $0.868 \scriptscriptstyle \pm \scriptstyle 8.3e\text{-}4$ & $0.823 \scriptscriptstyle \pm \scriptstyle 4.0e\text{-}4$ & $0.725 \scriptscriptstyle \pm \scriptstyle 9.5e\text{-}4$ & $5.657 \scriptscriptstyle \pm \scriptstyle 1.5e\text{-}2$ & $0.923 \scriptscriptstyle \pm \scriptstyle 1.0e\text{-}4$ & $0.973 \scriptscriptstyle \pm \scriptstyle 6.0e\text{-}4$ & $0.746 \scriptscriptstyle \pm \scriptstyle 6.0e\text{-}4$ \\
ResNet-T-LR     & $0.683 \scriptscriptstyle \pm \scriptstyle 6.1e\text{-}3$ & $0.862 \scriptscriptstyle \pm \scriptstyle 0.0e+00$ & $0.425 \scriptscriptstyle \pm \scriptstyle 7.4e\text{-}4$ & $3.030 \scriptscriptstyle \pm \scriptstyle 3.4e\text{-}2$ & $0.872 \scriptscriptstyle \pm \scriptstyle 7.3e\text{-}4$ & $0.822 \scriptscriptstyle \pm \scriptstyle 5.5e\text{-}4$ & $0.731 \scriptscriptstyle \pm \scriptstyle 1.1e\text{-}3$ & $5.471 \scriptscriptstyle \pm \scriptstyle 9.2e\text{-}3$ & $0.923 \scriptscriptstyle \pm \scriptstyle 5.8e\text{-}4$ & $0.975 \scriptscriptstyle \pm \scriptstyle 1.0e\text{-}4$ & $0.744 \scriptscriptstyle \pm \scriptstyle 3.3e\text{-}4$ \\
ResNet-P        & $0.675 \scriptscriptstyle \pm \scriptstyle 4.2e\text{-}3$ & $0.860 \scriptscriptstyle \pm \scriptstyle 6.2e\text{-}4$ & $0.453 \scriptscriptstyle \pm \scriptstyle 3.1e\text{-}3$ & $3.041 \scriptscriptstyle \pm \scriptstyle 1.7e\text{-}2$ & $0.872 \scriptscriptstyle \pm \scriptstyle 1.4e\text{-}3$ & $0.820 \scriptscriptstyle \pm \scriptstyle 2.0e\text{-}4$ & $0.733 \scriptscriptstyle \pm \scriptstyle 5.0e\text{-}4$ & $5.305 \scriptscriptstyle \pm \scriptstyle 2.3e\text{-}2$ & $0.923 \scriptscriptstyle \pm \scriptstyle 3.6e\text{-}4$ & $0.972 \scriptscriptstyle \pm \scriptstyle 2.1e\text{-}4$ & $0.744 \scriptscriptstyle \pm \scriptstyle 1.7e\text{-}4$ \\
ResNet-PLR      & $0.691 \scriptscriptstyle \pm \scriptstyle 6.3e\text{-}3$ & $0.861 \scriptscriptstyle \pm \scriptstyle 4.1e\text{-}4$ & $0.443 \scriptscriptstyle \pm \scriptstyle 1.4e\text{-}3$ & $3.040 \scriptscriptstyle \pm \scriptstyle 2.1e\text{-}2$ & $0.874 \scriptscriptstyle \pm \scriptstyle 5.0e\text{-}4$ & $0.825 \scriptscriptstyle \pm \scriptstyle 1.1e\text{-}3$ & $0.734 \scriptscriptstyle \pm \scriptstyle 6.3e\text{-}4$ & $5.400 \scriptscriptstyle \pm \scriptstyle 2.6e\text{-}2$ & $0.924 \scriptscriptstyle \pm \scriptstyle 2.9e\text{-}4$ & $0.975 \scriptscriptstyle \pm \scriptstyle 9.1e\text{-}5$ & $0.743 \scriptscriptstyle \pm \scriptstyle 4.0e\text{-}4$ \\
\midrule
Transformer-L   & $0.668 \scriptscriptstyle \pm \scriptstyle 1.3e\text{-}2$ & $0.861 \scriptscriptstyle \pm \scriptstyle 6.2e\text{-}4$ & $0.455 \scriptscriptstyle \pm \scriptstyle 1.4e\text{-}3$ & $3.188 \scriptscriptstyle \pm \scriptstyle 8.8e\text{-}3$ & $0.860 \scriptscriptstyle \pm \scriptstyle 6.5e\text{-}4$ & $0.824 \scriptscriptstyle \pm \scriptstyle 4.6e\text{-}4$ & $0.727 \scriptscriptstyle \pm \scriptstyle 1.1e\text{-}3$ & $5.434 \scriptscriptstyle \pm \scriptstyle 2.3e\text{-}2$ & $0.924 \scriptscriptstyle \pm \scriptstyle 1.1e\text{-}4$ & $0.973 \scriptscriptstyle \pm \scriptstyle 2.0e\text{-}4$ & $0.743 \scriptscriptstyle \pm \scriptstyle 2.7e\text{-}4$ \\
Transformer-LR  & $0.666 \scriptscriptstyle \pm \scriptstyle 1.0e\text{-}3$ & $0.861 \scriptscriptstyle \pm \scriptstyle 4.1e\text{-}4$ & $0.446 \scriptscriptstyle \pm \scriptstyle 1.1e\text{-}3$ & $3.193 \scriptscriptstyle \pm \scriptstyle 1.6e\text{-}2$ & $0.861 \scriptscriptstyle \pm \scriptstyle 2.0e\text{-}4$ & $0.824 \scriptscriptstyle \pm \scriptstyle 1.6e\text{-}3$ & $0.733 \scriptscriptstyle \pm \scriptstyle 7.8e\text{-}4$ & $5.430 \scriptscriptstyle \pm \scriptstyle 3.0e\text{-}2$ & $0.924 \scriptscriptstyle \pm \scriptstyle 1.8e\text{-}4$ & $0.973 \scriptscriptstyle \pm \scriptstyle 1.0e\text{-}4$ & $0.743 \scriptscriptstyle \pm \scriptstyle 1.8e\text{-}4$ \\
Transformer-Q-L & $0.704 \scriptscriptstyle \pm \scriptstyle 1.5e\text{-}3$ & $0.861 \scriptscriptstyle \pm \scriptstyle 1.1e\text{-}3$ & $0.426 \scriptscriptstyle \pm \scriptstyle 1.6e\text{-}3$ & $3.183 \scriptscriptstyle \pm \scriptstyle 2.5e\text{-}2$ & $0.869 \scriptscriptstyle \pm \scriptstyle 2.7e\text{-}4$ & $0.820 \scriptscriptstyle \pm \scriptstyle 3.1e\text{-}3$ & $0.735 \scriptscriptstyle \pm \scriptstyle 1.5e\text{-}3$ & $5.553 \scriptscriptstyle \pm \scriptstyle 1.5e\text{-}2$ & $0.925 \scriptscriptstyle \pm \scriptstyle 2.8e\text{-}4$ & $0.976 \scriptscriptstyle \pm \scriptstyle 5.9e\text{-}5$ & $0.744 \scriptscriptstyle \pm \scriptstyle 2.0e\text{-}4$ \\
Transformer-Q-LR & $0.690 \scriptscriptstyle \pm \scriptstyle 1.9e\text{-}3$ & $0.857 \scriptscriptstyle \pm \scriptstyle 2.4e\text{-}4$ & $0.425 \scriptscriptstyle \pm \scriptstyle 1.2e\text{-}3$ & $3.143 \scriptscriptstyle \pm \scriptstyle 1.6e\text{-}2$ & $0.868 \scriptscriptstyle \pm \scriptstyle 4.9e\text{-}4$ & $0.818 \scriptscriptstyle \pm \scriptstyle 2.3e\text{-}3$ & $0.726 \scriptscriptstyle \pm \scriptstyle 1.2e\text{-}3$ & $5.471 \scriptscriptstyle \pm \scriptstyle 1.5e\text{-}2$ & $0.924 \scriptscriptstyle \pm \scriptstyle 2.0e\text{-}4$ & $0.975 \scriptscriptstyle \pm \scriptstyle 1.9e\text{-}4$ & $0.744 \scriptscriptstyle \pm \scriptstyle 3.5e\text{-}4$ \\
Transformer-T-L & $0.693 \scriptscriptstyle \pm \scriptstyle 6.8e\text{-}3$ & $0.862 \scriptscriptstyle \pm \scriptstyle 2.4e\text{-}4$ & $0.439 \scriptscriptstyle \pm \scriptstyle 1.0e\text{-}3$ & $3.136 \scriptscriptstyle \pm \scriptstyle 3.5e\text{-}3$ & $0.872 \scriptscriptstyle \pm \scriptstyle 1.3e\text{-}4$ & $0.826 \scriptscriptstyle \pm \scriptstyle 2.3e\text{-}3$ & $0.731 \scriptscriptstyle \pm \scriptstyle 1.6e\text{-}3$ & $5.579 \scriptscriptstyle \pm \scriptstyle 5.2e\text{-}2$ & $0.924 \scriptscriptstyle \pm \scriptstyle 4.0e\text{-}4$ & $0.977 \scriptscriptstyle \pm \scriptstyle 2.1e\text{-}4$ & $0.743 \scriptscriptstyle \pm \scriptstyle 2.3e\text{-}4$ \\
Transformer-T-LR & $0.686 \scriptscriptstyle \pm \scriptstyle 4.1e\text{-}3$ & $0.862 \scriptscriptstyle \pm \scriptstyle 1.1e\text{-}3$ & $0.423 \scriptscriptstyle \pm \scriptstyle 3.4e\text{-}3$ & $3.149 \scriptscriptstyle \pm \scriptstyle 1.4e\text{-}2$ & $0.871 \scriptscriptstyle \pm \scriptstyle 8.0e\text{-}4$ & $0.823 \scriptscriptstyle \pm \scriptstyle 2.4e\text{-}3$ & $0.733 \scriptscriptstyle \pm \scriptstyle 9.4e\text{-}4$ & $5.515 \scriptscriptstyle \pm \scriptstyle 2.0e\text{-}2$ & $0.924 \scriptscriptstyle \pm \scriptstyle 6.1e\text{-}5$ & $0.976 \scriptscriptstyle \pm \scriptstyle 9.2e\text{-}5$ & $0.744 \scriptscriptstyle \pm \scriptstyle 2.9e\text{-}4$ \\
Transformer-PLR & $0.686 \scriptscriptstyle \pm \scriptstyle 6.2e\text{-}3$ & $0.864 \scriptscriptstyle \pm \scriptstyle 9.4e\text{-}4$ & $0.449 \scriptscriptstyle \pm \scriptstyle 1.2e\text{-}3$ & $3.091 \scriptscriptstyle \pm \scriptstyle 1.3e\text{-}2$ & $0.873 \scriptscriptstyle \pm \scriptstyle 1.5e\text{-}3$ & $0.823 \scriptscriptstyle \pm \scriptstyle 1.7e\text{-}3$ & $0.734 \scriptscriptstyle \pm \scriptstyle 2.1e\text{-}4$ & $5.581 \scriptscriptstyle \pm \scriptstyle 6.4e\text{-}2$ & $0.924 \scriptscriptstyle \pm \scriptstyle 1.8e\text{-}4$ & $0.975 \scriptscriptstyle \pm \scriptstyle 2.2e\text{-}4$ & $0.743 \scriptscriptstyle \pm \scriptstyle 2.4e\text{-}4$ \\
\bottomrule
\end{tabular}}
\end{sidewaystable}

\end{document}